\documentclass[lettersize,journal]{IEEEtran}
\usepackage{amsmath,amsfonts}
\usepackage{algorithmic}
\usepackage{algorithm}
\usepackage{array}
\usepackage[caption=false,font=normalsize,labelfont=sf,textfont=sf]{subfig}
\usepackage{textcomp}
\usepackage{stfloats}
\usepackage{url}
\usepackage{verbatim}
\usepackage{graphicx}
\usepackage{cite}
\usepackage{threeparttable}
 
\DeclareMathAlphabet{\mathpzc}{OT1}{pzc}{m}{it} 
\usepackage{multirow} 
\usepackage{color,soul} 
\soulregister{\cite}7 
\soulregister{\citep}7 
\soulregister{\citet}7 
\soulregister{\ref}7 
\soulregister{\pageref}7 
\newcommand{\mathcolorbox}[2]{\colorbox{#1}{$ #2$}}  

\sethlcolor{yellow} 
\setstcolor{green} 
\setulcolor{red} 

\begin{document}

\title{A Temporal Densely Connected Recurrent Network for Event-based Human Pose Estimation}

\author{Zhanpeng Shao,~\IEEEmembership{Member, ~IEEE,}
        Wen Zhou,
        Wuzhen Wang,
        Jianyu Yang, ~\IEEEmembership{Member, ~IEEE,}
        Youfu Li,~\IEEEmembership{~Fellow, IEEE}
\thanks{This work was supported in part by the National Natural Science Foundation of China under Grant 61976191, Grant 62203168, and Grant 61603341, in part by the Zhejiang Provincial Natural Science Foundation under Grant LY19F030015. 
}
\thanks{Zhanpeng Shao is with the College of Information Science and Engineering, Hunan Normal University, 36 Lushan Road, Changsha, China (e-mail: zpshao@hunnu.edu.cn).}
\thanks{Wen Zhou and Wuzhen Wang is with the Department of Computer Science and Technology, Zhejiang University of Technology, 288 Liuhe Road, Hangzhou, China (e-mail: \{wenzhou, kuretru\}@zjut.edu.cn).}
\thanks{Youfu Li is with the Department of Mechanical Engineering, City University of Hong Kong, 83 Tat Chee Avenue, Hong Kong (email: meyfli@cityu.edu.hk).}
\thanks{Jianyu Yang is with the School of Rail Transportation, Soochow University, 8 Jixue Road, Suzhou, China (e-mail: jyyang@suda.edu.cn).}}

\markboth{Journal of \LaTeX\ Class Files,~Vol.~14, No.~8, August~2021}%
{Shell \MakeLowercase{\textit{et al.}}: A Sample Article Using IEEEtran.cls for IEEE Journals}

\IEEEpubid{0000--0000/00\$00.00~\copyright~2021 IEEE}

\maketitle

\begin{abstract}
Event camera is an emerging bio-inspired vision sensors that report per-pixel brightness changes asynchronously. It holds noticeable advantage of high dynamic range, high speed response, and low power budget that enable it to best capture local motions in uncontrolled environments. This motivates us to unlock the potential of event cameras for human pose estimation, as the human pose estimation with event cameras is rarely explored. Due to the novel paradigm shift from conventional frame-based cameras, however, event signals in a time interval contain very limited information, as event cameras can only capture the moving body parts and ignores those static body parts, resulting in some parts to be incomplete or even disappeared in the time interval. This paper proposes a novel densely connected recurrent architecture to address the problem of incomplete information. By this recurrent architecture, we can explicitly model not only the sequential but also non-sequential geometric consistency across time steps to accumulate information from previous frames to recover the entire human bodies, achieving a stable and accurate human pose estimation from event data. Moreover, to better evaluate our model, we collect a large scale multimodal event-based dataset that comes with human pose annotations, which is by far the most challenging one to the best of our knowledge. The experimental results on two public datasets and our own dataset demonstrate the effectiveness and strength of our approach. Code \footnote{\url{https://github.com/xavier-zw/tDenseRNN_pose}} can be available online for facilitating the future research.
\end{abstract}


\section{Introduction}
\label{sec:intro}

\begin{figure}[ht]
\centering

\includegraphics[width=3.5in]{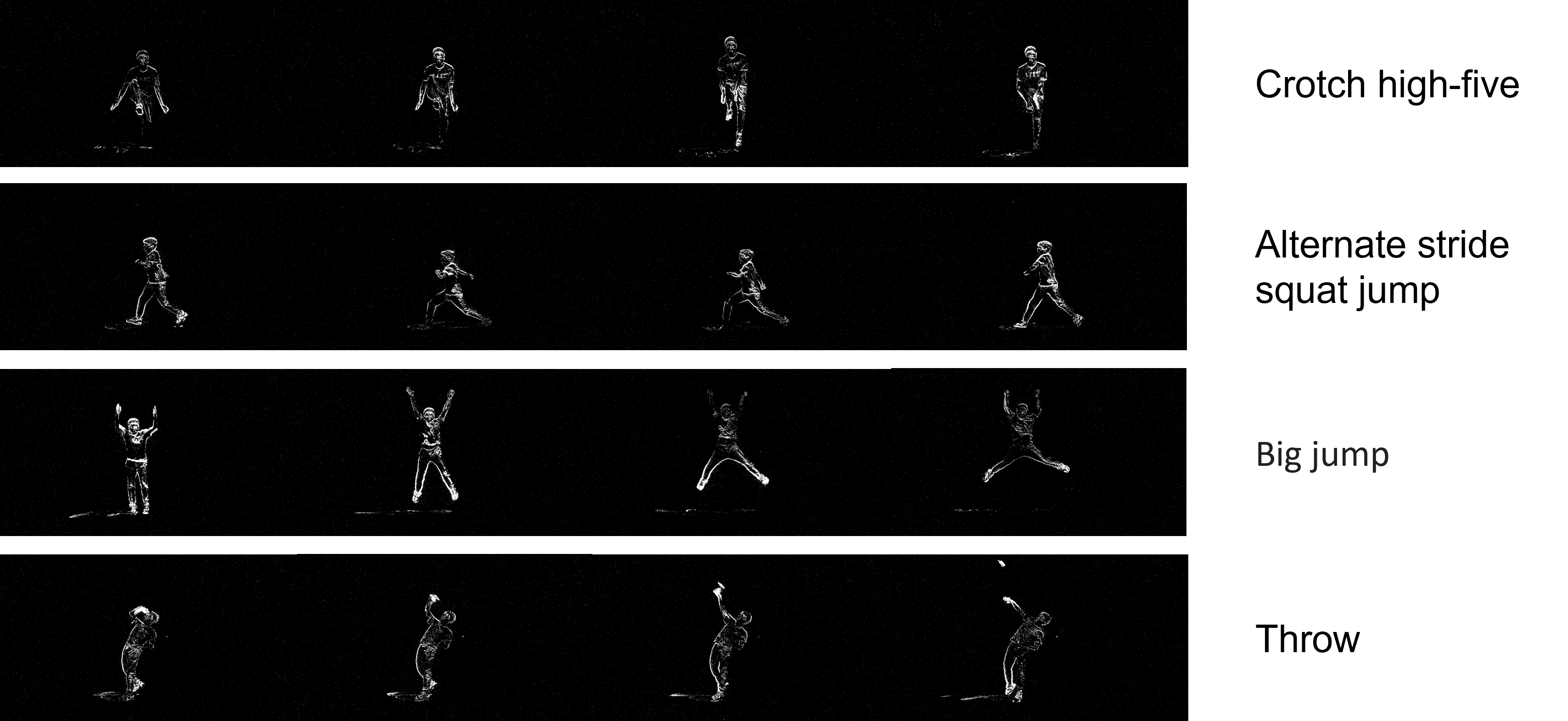}

\caption{Action samples captured by an event camera.  The human body parts are not all visible during movements, since that some body parts are just static so that no events are triggered.}
\label{events_fig}
\end{figure}

Human pose estimation is a fundamental vision task which finds a variety of real applications in computer vision, such as augmented reality, human behavior understanding, and human-robot interaction. Existing efforts \cite{xiao2018simple,chen2018cascaded,Cheng_2020_CVPR,Openpose2021} are mainly based on conventional RGB cameras, and meanwhile remarkable advances have been made thanks to the success of deep CNNs. Meanwhile, event cameras \cite{Guillermo2020,patrick2008128x}, inspired by the biological vision process, output a stream of asynchronous per-pixel brightness changes, which allows event cameras to best detect and capture local motions.
Event camera has recently been an emerging vision sensor to offer opportunities for computer vision applications \cite{calabrese2019dhp19,Guillermo2020,HuangCeleX,Maqueda2018}, such as SLAM\cite{Elias2017}, gesture recognition \cite{Arnon2017,wang2019ev}, optical flow estimation \cite{Zhu-RSS-18}, object recognition \cite{sironi2018hats,deng2021mvf,GarrickHfirst2015} due to their unique advantages of very high temporal resolution, high dynamic range, low latency, and low power budget, which can tackle the problems that are difficult with the conventional cameras. Therefore, event camera has a large potential for human pose estimation which may operate under uncontrolled lighting conditions and crowded backgrounds. However, the potential of event cameras in human pose estimation is being under-explored. 



\IEEEpubidadjcol 
It is straightforward to apply the existing approaches based on conventional cameras \cite{xiao2018simple,chen2018cascaded,Cheng_2020_CVPR} to the event-based pose estimation by accumulating asynchronous events into frames \cite{calabrese2019dhp19,Maqueda2018}. Nevertheless, the sparsity of event data limits the performance, because event cameras only output events at those positions that have noticeable brightness changes in the scene. For example, in \figurename~\ref{events_fig} we list several human action samples captured by event cameras, where the human body is partial invisible in some frames, because event cameras can only capture the moving body parts and ignores those static body parts, resulting in some parts to be incomplete or even disappeared. Therefore, this problem motivates us to consider the geometric consistency and temporal dependency across frames for event-based pose estimation to help recover and complete those lost information. Existing approaches for video-based human pose estimation \cite{song2017thin,luo2018lstm,zou2021eventhpe,nie2019dynamic,li2020exploring} can be leveraged to address this. However, it is not always effective for event-based pose estimation, because in some cases no frame contains a complete human body whose all body parts are visible in an event video, as shown in \figurename~\ref{events_fig}. As a result, the existing approaches that consider geometric consistency between adjacent frames may not be useful to help complete invisible body parts.

In this paper, we argue that the temporal geometric consistency in event videos may require not only a sequential temporal dependency but also a non-sequential information accumulation across frames in a long range to recover the whole human bodies. For instance, as shown in \figurename~\ref{events_fig}, in some cases there are some key frames that contains all body parts, which can be used to complete the information of neighboring frames. In some cases, we cannot always have key frames in a video so that we need to accumulate the information from a set of frames in a time window to recover the full human body. Therefore, we adopt a basic recurrent architecture with a newly proposed temporal dense connection across a sequence of time steps to capture the geometric consistency of human poses across frames in both local and long range to complete the lost information in event frames, as illustrated in \figurename~\ref{overview_fig}. Specifically, we incorporate a set of dense connections between the current frame and its all preceding frames into the recurrent network built by using a Long Short-Term Memory (LSTM) module to link an encoder-decoder CNN \cite{xiao2018simple} temporally. This new architecture allows for both the sequential and non-sequential temporal dependency modeling thanks to such skipped dense connections rather than only sequential connections between two neighboring frames in \cite{luo2018lstm,nie2019dynamic,artacho2020unipose,dang2021relation}. Moreover, we introduce a spatio-temporal attention mechanism into the dense connections to pay different importance to the preceding frames and their spatial joints when fusing their information to the current frame. In addition, we found in the experiments that the existing event-based human pose datasets \cite{calabrese2019dhp19,zou2021eventhpe} are normally captured under indoor environments with clean background and controlled lighting conditions. Therefore, our method is easily saturated in the performance with these datasets. Therefore, to evaluate our method, we collect a new event-based human pose dataset, referred as CDEHP, to provide the benchmarks for event-based pose estimation.

The contributions of this paper can be summarized as following: 
\begin{itemize} 
\item We present a novel recurrent architecture with a newly proposed temporal dense connections to capture both sequential and non-sequential geometric consistency and dependency among event frames for event-based human pose estimation. Our method can effectively overcome the problem of incomplete event information in event frames. 
\item The spatio-temporal attention mechanism is introduced in our method to provide a more effective information fusion and completion for event data by paying different importance to preceding frames and their spatial joints, compared with simply accumulating them. 
\item We collect a large scale multimodal event-based human pose dataset from Color (RGB), Depth, and Event cameras, called as CDEHP in this paper. The dataset is captured from outdoor environments under varying light conditions, while most existing event-based pose datasets are captured from controlled indoor environments. The CDEHP can help unlock the potential of event cameras in human pose estimation, and facilitate the existing and future related research directions.
\end{itemize}

\section{Related Work}
\label{sec:related}
\subsection{Human Pose Estimation}
Human pose estimation from still images in early works usually starts from building the parts-based graphic models or pictorial structure models \cite{felzenszwalb2005pictorial,andriluka2009pictorial,pishchulin2013poselet,yang2012articulated} to learn spatial relationships between articulated body parts. Recently, the performance of these earlier works have been surpassed largely thanks to the great success of deep convolutional networks \cite{newell2016stacked,luo2018lstm,Cheng_2020_CVPR,xiao2018simple,Openpose2021,insafutdinov2016deepercut,chen2018cascaded,shao2021icra} that provide dominant solutions nowadays. Apart from image-based pose estimation, many efforts \cite{luo2018lstm,song2017thin,pfister2015flowing,girdhar2018detect} has also been made to exploit temporal and motion information for human pose estimation from videos by using optical flow or 3D CNNs, which are related to our work since that we consider to accumulate event signals into a sequence of event frames. However, these methods have the limited ability to extract temporal contexts explicitly. More recently, recurrent architectures \cite{luo2018lstm,li2020exploring,nie2019dynamic,dang2021relation,artacho2020unipose,Liu_2021_CVPR} is normally integrated with an encoder-decoder CNN framework to model the temporal dependency across frames to refine pose predictions, which have been pioneering frameworks for video-base human pose estimation. Such kind of frameworks share a general structure, where CNNs are normally used to encode and decode every frame sequentially, and a recurrent mechanism is then introduced to temporally link the encoder-decoder streams along time steps to propagate temporal dynamics between neighboring frames. However, they all just model the temporal dependency between two consecutive frames, which are not always effective for human pose estimation from event signals, because in an event video we also need to consider a long-range geometric consistency across a set of frames in a time window. DCPose \cite{Liu_2021_CVPR} leverages the temporal cues between past, current, and next frames to facilitate keypoint prediction. However, it still models a short-range temporal dependency, and meanwhile it has to depends on future frames. Similarly, most recent work FAMI-Pose \cite{liu2022temporal} leverages a hierarchical alignment framework to update a short-range of neighboring frames (e.g., 2 previous and 2 future frames) to align with the current frame at feature level.



Different from existing methods, our model introduces dense connections across a set of consecutive frames rather than one single temporal connection between the current and last frames to help recover human poses in event streams.


\subsection{Event Camera and Applications}
\begin{figure}[ht]
\centering
\includegraphics[width=2.6in]{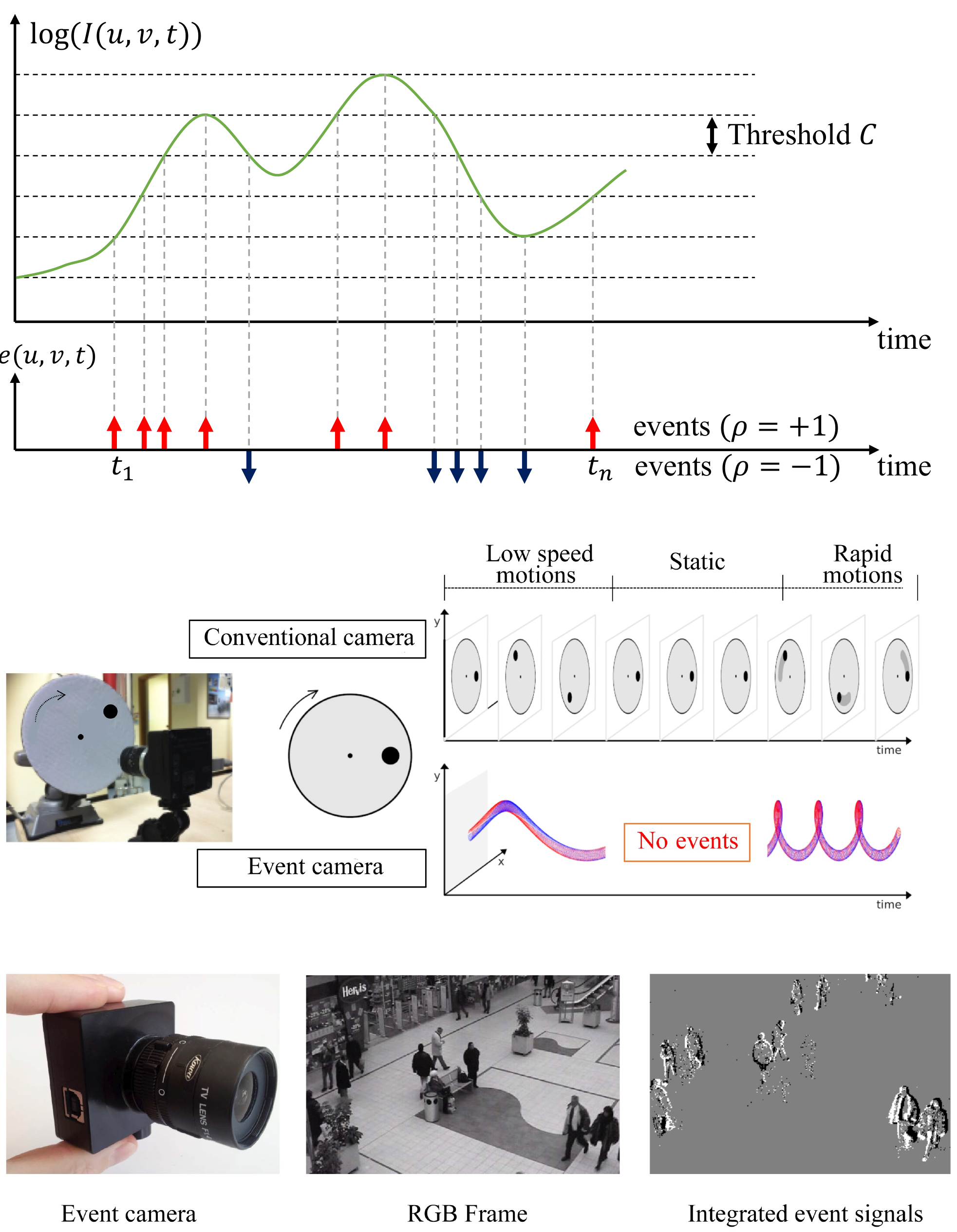}
\caption{\textbf{Top:} Event-based vision sensor working principle. Events are generated asynchronously according to the brightness changes ($\Delta log(I)$). Dotted lines show how the thresholds for detecting increases and decreases in brightness change as outputs are generated. This sub-figure is cropped from \cite{GarrickHfirst2015}.} \textbf{Middle:} The difference in vision acquisition between the event camera and RGB camera. \textbf{Bottom:} The RGB frame captured from the conventional RGB camera and event frames accumulated from event signals. Best viewed in color.
\label{event_camera_fig}
\end{figure}

Most existing methods and algorithms in vision research are based on the conventional frame-based cameras, where the cameras output a sequence of synchronous frames at a fixed rate. The frame contains the entire information in scene, thereby producing redundant vision signals especially when there are only very little changes in the scene between consecutive frames. In contrary, event cameras \cite{patrick2008128x,Guillermo2020}, a new type of bio-inspired vision sensors that have emerged in the last few years, output a signed ‘event’ signal $e_i$ at the pixel location of $(u_i,v_i)$ only when the detected changes in the log brightness exceeds a predefined threshold $C$ $(\left | \Delta log(I(u_i,v_i,t_i)) \right |>C)$ at time instance $t_i$ with the polarity $\rho_i$ $(\rho \in \{-1,+1\})$. Then, an event signal can be represented as $e_i=(u_i,v_i,t_i,\rho_i)$, where positive events $(\rho_i=+1)$ indicates the brightness increase $(\Delta log(I) >C)$ and negative events $(\rho_i=-1)$ indicates the brightness decrease $(\Delta log(I) <-C)$, as shown in \figurename~\ref{event_camera_fig}.


Thus, unlike conventional cameras, event cameras produce a sequence of asynchronous events because they asynchronously sample light of each pixel independently. As a result, the events are spatially much sparser in comparison with conventional frame-based cameras, where each frame is generated by densely sampling the entire pixels at the same time. Hence, event camera is able to best capture local motions in the scene as a stream of sparse and asynchronous events. Event cameras have unique advantages of very high temporal resolution, high dynamic range, low latency, and low power consumption. Hence, event cameras have stimulated a variety of research activities and applications in computer vision \cite{Guillermo2020}, including visual SLAM \cite{Elias2017}, optical flow estimation \cite{Zhu-RSS-18}, object tracking \cite{HuangCeleX}, object recognition \cite{sironi2018hats,deng2021mvf,Deng2021distillation}, gait recognition \cite{wang2019ev}, and high-speed maneuvers \cite{MuegglerHighspeed}, among others.


For human pose estimation, there has been little investigations into applying event cameras to capture and estimate human poses. DHP19 \cite{calabrese2019dhp19}, EventCap \cite{Xu2020EventCapM3}, and EventHPE \cite{zou2021eventhpe} are perhaps most related to our work. In DHP19 \cite{calabrese2019dhp19}, a simple encoder-decoder CNN is designed to estimate human poses from single integrated event frames. The method normally fails in those frames that contain only few visible body parts. Other two related works EventCap and EventHPE are mainly dedicated to address 3D human motion capture and shape recovery rather than pose estimation from event streams. In EventCap \cite{Xu2020EventCapM3}, it aims to achieve markerless human motion capture, which however requires a necessary additional input of gray-scale images apart from event signals. EventHPE \cite{zou2021eventhpe} can achieve human pose and shape estimation simultaneously with the solo event source. However, it needs to fuse an event sequence and its corresponding optical flows, where optical flows require an extra FlowNet to infer. Moreover, a beginning body posture at the first frame of gray-scale images is necessary for this method. Compared with them, our approach only takes in event signals as the sole input source to address the human pose estimation with a light model.

\subsection{Event-based Human Pose Datasets}
There are already many RGB-based datasets for human pose estimation, including images and videos, such as COCO \cite{Lin2014}, MPII \cite{MPII2014}, LSP \cite{LSP2011}, PoseTrack \cite{Iqbal2017PoseTrackJM}, etc., which have largely facilitated the evaluations of RGB-based algorithms. However, event-based human pose datasets have been rarely curated. There are only two existing datasets, DHP19 \cite{calabrese2019dhp19} and MMHPSD \cite{zou2021eventhpe}, which are commonly used to evaluate the pose estimation. DHP19 contains total of 33 action categories, but most of them are simple slow motions such as hand waving and leg swinging so that a very simple CNN can even saturate on this dataset. The MMHPSD dataset is a multimodal dataset that contains more diverse actions with fast, medium, and slow motions. However, the 2D joint annotations are roughly obtained by using OpenPose \cite{Openpose2021} as MMHPSD dataset was initially created for 3D human shape estimation. Moreover, both datasets were acquired indoors under a fixed environment with clean backgrounds. Therefore, we curated a large-scale multimodal event-based dataset CDEHP that are captured in outdoor environments with uncontrolled lighting conditions.



\begin{figure*}[ht]
\centering
\includegraphics[width=7in]{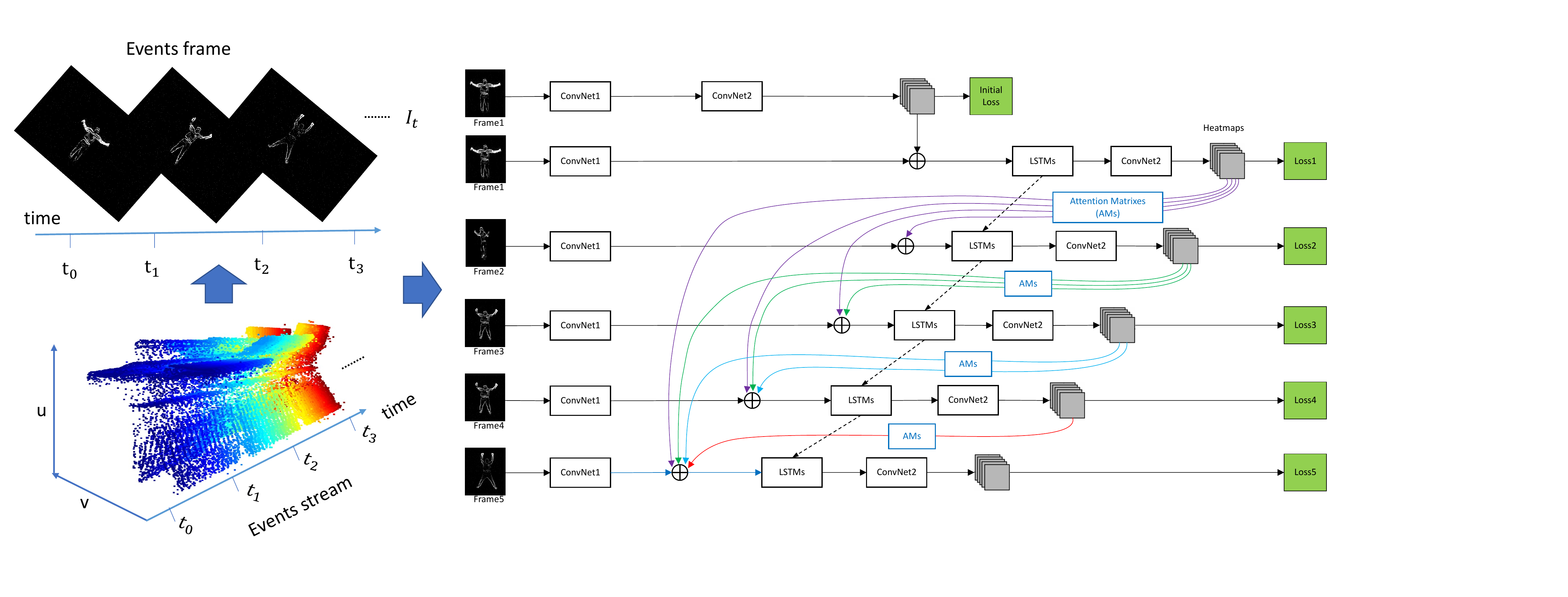}
\caption{An overview of the densely connected recurrent network. Events in a fixed time interval ($t_{i-1},t_i$) are accumulated as an event frame first. We then can convert a stream of asynchronous event signals to an event video that consists of a sequence of event frames $I_t$. \textit{ConvNet1} is an encoder-decoder CNN to extract convolutional features from event frames, and \textit{ConvNet2} that has a $1\times1$ convolutional layer is added to generate the heatmaps. An LSTM module is incorporated with \textit{ConvNet1} and \textit{ConvNet2} to create a basic recurrent network, based on which we introduce a set of dense connections with attentions to pass all the preceding heatmaps to fused with convolutional features at the current time step (frame) for explicitly modeling both the sequential and non-sequential geometric consistency across time steps in a long range.}
\label{overview_fig}
\end{figure*}

\section{Our Approach}
\label{sec:approach}
In the event capturing, event camera provides a stream of asynchronous event signals. In this work, we divide the event stream into a sequence of $T$ event packets, where each event packet consists of a set of event signals collected from a fixed time interval $(t_{i-1},t_i)$, as shown in \figurename~\ref{overview_fig}. We then accumulate the events of each packet by following the strategy in \cite{Maqueda2018} to integrate an event frame. We here have a little abuse of notation $t$ that is used to represent the temporal index of integrated event frames rather than a time instance. Therefore, we address the event-based human pose estimation by detecting the keypoints from a sequence of $T$ consecutive event frames $\{I_t\}_{t=0}^{T-1}$, i.e., $I \in \mathbb{R}^{W \times H \times T}$. Most of the existing methods transform this problem to predict a set of heatmaps $\mathbf{b}=\{\mathbf{b}_t\}_{t=0}^{T-1}$ for all frames. The $\mathbf{b}_t \in \mathbb{R}^{W' \times H' \times K}$ is of spatial size $W' \times H'$, where $K$ represents the number of keypoints in a human body, and each $\mathbf{b}_t(k)$ indicates the location confidence of the $k$-th keypoint at the $t$-th time step (frame).

\figurename~\ref{overview_fig} illustrates an overview of our proposed recurrent architecture. A base convolutional network based on the encoder-decoder architecture \cite{xiao2018simple} (i.e.,\textit{ConvNet1} in \figurename~\ref{overview_fig}) are temporally linked by using a recurrent architecture for event-based pose estimation, where specifically a convolutional LSTM module is employed to model the temporal dependency across consecutive frames. At every time steps, \textit{ConvNet1} first learns the convolutional features and then recovers the high-resolution spatial feature maps for heatmap generation. These feature maps across time steps are then sent into the LSTM module to form a recurrent network. A following $1\times1$ convolutional network (i.e.,\textit{ConvNet2} in \figurename~\ref{overview_fig}) finally produces the heatmaps of joint locations for each input frame. Most importantly, to address the problem of event-based pose estimation, we introduce a set of temporal dense connections into such recurrent network inspired by prior works \cite{Huangdense,luo2018lstm}, where a spatio-temporal attention mechanism is further incorporated to learn the their weights of connections.

\subsection{Convolutional Base Network}
We follow the widely-used pipelines \cite{newell2016stacked,xiao2018simple,sun2019deep,li2020exploring} to predict the body keypoints by using an encoder-decoder architecture, which usually consists of a series of convolutional layers followed by a couple of upsampling and deconvolutional layers to generate high-resolution spatial feature maps. A few of regression layers take the feature maps as input to finally estimate the heatmaps for all keypoints. We adopt the convolutional network in \cite{xiao2018simple}, i.e., SimpleBaseline, as our base network to extract the feature maps for each frame to generate heatmaps, as the SimpleBaseline \cite{xiao2018simple} is a very simple but effective convolutional network for human pose estimation from still images.  

SimpleBaseline simply uses the ResNet \cite{he2016deep} as the backbone to extract high-level image feature and then typically adds three transposed convolution layers (deconvolution) to recover the high-resolution spatial feature maps. Specifically, there are $256$ filters with a $4 \times 4$ kernel and a stride of 2 in each deconvolutional layer. An $1 \times 1$ convolutional layer is finally adopted at last to predict the heatmaps of the body keypoints based on the feature maps. In our architecture, we use the abbreviation \textit{ConvNet2} to indicate such an $1 \times 1$ convolutional layer, while we use the abbreviation \textit{ConvNet1} to indicate the convolutional block composed of the ResNet and three deconvolution layers.

\subsection{LSTM-based Recurrent Network}
To build our recurrent network, we use an LSTM module to link the convolutional base network in temporal dimension, as illustrated in \figurename~\ref{overview_fig}, where we choose to insert the LSTM module between \textit{ConvNet1} and \textit{ConvNet2}. By such recurrent style model, the parameters of \textit{ConvNet1} and \textit{ConvNet2} are shared across different time steps. Since the LSTM module take the convolutional feature maps generated by the \textit{ConvNet1} as input, we leverage two convolutional LSTM layers \cite{Shi2015ConvolutionalLN} without peephole connections as our LSTM module, 
where the operations inside it are defined as: 
\begin{equation}
\begin{aligned}
&i_t = \delta \left (\mathbf{W}_{xi}*X_t+ \mathbf{W}_{hi}*h_{t-1}+\epsilon_i\right ), \\
&f_t = \delta \left (\mathbf{W}_{xf}*X_t+ \mathbf{W}_{hf}*h_{t-1}+\epsilon_f\right ), \\
&o_t = \delta \left (\mathbf{W}_{xo}*X_t+ \mathbf{W}_{ho}*h_{t-1}+\epsilon_o\right ), \\
&C_t = f_t \odot C_{t-1} + i_t \odot \varphi \left (\mathbf{W}_{xc}*X_t+ \mathbf{W}_{hc}*h_{t-1}+\epsilon_c\right ), \\ 
&h_t = o_t \odot  \varphi(C_t),  \\
\end{aligned}
\label{eq1}
\end{equation}
Unlike the standard LSTM, '*' here denotes a convolutional operation similar with \cite{Shi2015ConvolutionalLN}. As a result, all the '+' in Eq.~\ref{eq1} represent the element-wise addition. $X_t$ denotes the input feature, while ${\bf{W}}_{**}$ denotes the weights of the LSTM module. $\epsilon_{*}$ denote the bias terms. $\delta$ and $\varphi$ represent the activation functions $sigmoid(\cdot)$ and $tanh(\cdot)$, respectively. $i_t$, $f_t$, $o_t$, and $h_t$ represents the input gate, forget gate, output gate, and hidden state, respectively. 
Such convolutional design of LSTM gates allows for more attentions on the regional context rather than global information, and it can effectively capture the local changes of joints.

We adopt two such convolutional LSTM layers with $3 \times 3$ kernels as our LSTM module, referred to as the function $\mathpzc{L\left(\cdot \right)}$. Successive event frames in an event video are sent into the feature extractor \textit{ConvNet1} to produce the feature maps of size $W' \times H' \times M$ as the input of the LSTM module across each time step. Lastly, the outputs of the LSTM module are sent into the heatmap generator \textit{ConvNet2} with an $1 \times 1$ convolutional layer to predict the heatmaps for each frame. Mathematically, we denote the computation of \textit{ConvNet1} and \textit{ConvNet2} consistently used in all time steps as the functions $\mathpzc{F}\left( \cdot \right)$ and $\mathpzc{g}\left( \cdot \right)$, respectively, and $I_t \in \mathbb{R}^{W \times H} $ is the original input event frame at time step $t$. Denote $\mathbf{b}_t \in \mathbb{R}^{W' \times H' \times K}$ as the heatmaps in time step $t$. The LSTM-based recurrent network can be formulated mathematically as following, 
\begin{align}
\mathbf{b}_t = \mathpzc{g}\left( \mathpzc{L} \left( \mathpzc{F} \left( I_t\right)\right) \right), t=0,1,2,...,T-1
\label{eq2}
\end{align}
\subsection{Dense Connections with Attentions}
The key strategy in this paper is that we propose a novel densely connectivity pattern as shown in \figurename~\ref{overview_fig} in the recurrent network to pass all the preceding heatmaps to the next time step for explicitly building sequential and non-sequential geometric consistency in an event video. We introduce direct connections from the heatmaps in every time step to the inputs of LSTM modules in all subsequent time steps. In this way, the $t$-th LSTM module $\mathpzc{L\left(\cdot \right)}$ receives the heatmaps in all preceding time steps. All these past heatmaps are simply accumulated as one heatmap, which is then concatenated (indicated by the operation $\oplus$) with the feature map extracted by \textit{ConvNet1} $\mathpzc{F}\left( \cdot \right)$ in the $t-th$ step as input to the LSTM module. Mathematically, a new recurrent architecture with the proposed temporal dense connections can be formulated as:
\begin{align}
\mathbf{b}_t = \mathpzc{g}\left( \mathpzc{L}\left( \mathpzc{F} \left( I_t\right) \oplus \sum_{\tau=0}^{t-1}\mathbf{b}_{\tau } \right) \right),
\label{eq3}
\end{align}
The preceding frames may play different important roles in predicting poses at current frame. Meanwhile, the visible body parts in the previous frames also may make different contributions on completing the lost body parts in the current frames. Therefore, we introduce an attention weight matrix into the fusion of the preceding heatmaps during the propagation. In addition, it should be noted that there are not prior heatmaps when $t=0$. Thus, we have to employ the base network (\textit{ConvNet1 + ConvNet2}) to generate the heatmap at the first frame to provide initialized prior input to LSTM module for the first frame self, as illustrated in \figurename~\ref{overview_fig}. Consequently, we have:
\begin{align}
&\mathbf{b}_t = \mathpzc{g}\left( \mathpzc{L}\left( \mathpzc{F} \left( I_t\right) \oplus \mathpzc{g}\left(   \mathpzc{F} \left( I_t\right)\right) \right) \right), t=0,
\label{eq4}
\end{align}
\begin{align}
&\mathcolorbox{yellow}{\mathbf{b}_t= \mathpzc{g}\left( \mathpzc{L}\left( \mathpzc{F} \left( I_t\right) \oplus \sum_{\tau=0}^{t-1} \left( \mathbf{W}^{t}_{\tau} \odot \mathbf{b}_{\tau } \right) \right) \right), t=1,...,T,
}
\label{eq5}
\end{align}

where $\mathbf{b}_t \in \mathbb{R}^{W' \times H' \times K}$, and $\mathbf{W}^{t}_{\tau}$ is the attention matrix to weight the heatmap of the $\tau$-th frame to fuse with the feature maps at $t$-th frame. $\odot$ indicates the element-wise multiplication that means the spatial attention on different body parts in the corresponding frame is also applied. Such spatio-temporal attention mechanism is introduced into the dense connections to pay different importance to the preceding frames and their spatial joints when fusing their information to the current frame. Because of such temporal connectivity, we refer to this recurrent architecture as Temporal Densely Connected Recurrent Network (tDenseRNN). This new architecture allows for both the sequential and non-sequential temporal dependency modeling thanks to such skipped dense connections rather than only sequential connections between two consecutive frames in \cite{luo2018lstm,nie2019dynamic,artacho2020unipose}. 
\subsection{Training of the Network}
For the ground truth heatmap for $k$-th joint, we transform the label in Cartesian coordinates for the $k$-th joint into a heatmap by placing a Gaussian peak at the center of the joint locations. We have $T$ time steps in the recurrent network, where $T$ is the number of consecutive frames in each training sequence. The total loss is accumulated at the end of each time step to supervise the network training. In the training, we aim to minimize such loss with the $l_2$ distance between the predicted heatmaps and ground truth heatmaps for all joints and all frames jointly, 
\begin{align}
l= \sum_{t}^{T}\sum_{k}^{K}\left \| \mathbf{b}_t(k)-\mathbf{b}_{t}^{*}(k) \right \|,
\label{eq6}
\end{align}

where $\mathbf{b}_t(k)$ and $\mathbf{b}_{t}^{*}(k)$ are the predicted heatmaps and ground truth heatmaps for $k$-th joint at $t$-th time step, respectively. 

\subsection{Our CDEHP Dataset}
\label{sec:cedhp}
A multimodal human pose dataset captured in outdoor scenes, CDEHP, has been curated to evaluate our approach, because there is not a very challenging dataset captured in different environments to facilitate the evaluation of our approach which easily saturates on the existing datasets with simple actions captured in a fixed indoor environment. 

\textbf{Data acquisition}. The CDEHP dataset was collected from different outdoor environments under varying light conditions by using a multiple camera system. The multiple camera system contains 2 different imaging sensors, including one event camera CeleX-V \cite{HuangCeleX} and one RGB-D camera Intel RealSense D435i, which could output a sequence of event streams, RGB color frames, and depth frames simultaneously. Specifically, the resolution of RGB-D camera (Intel RealSense D435i) is set to be 840*480 with the frame rate of 60, while the resolution of the event camera (Celex-V) is set to be 1280*800. The samples in the CDEHP dataset are captured in four different outdoor environments by 20 subjects, with 15 being male and 5 being female. Each subject performs 25 different actions with varying speeds (slow/medium/fast) for $3\sim4$ times, as listed in \tablename~\ref{action_list}. Finally, this amounts to 500 video samples collected in total, where each video sample contains RGB video sequences, depth video sequences, and event streams. Overall, this amounts to be a total of 82K frames in the dataset in terms of the number of RGB-D frames. Some action samples with RGB and event modalities are shown in a figure of Appendix.

\begin{figure}[ht]
\centering
\includegraphics[width=2in]{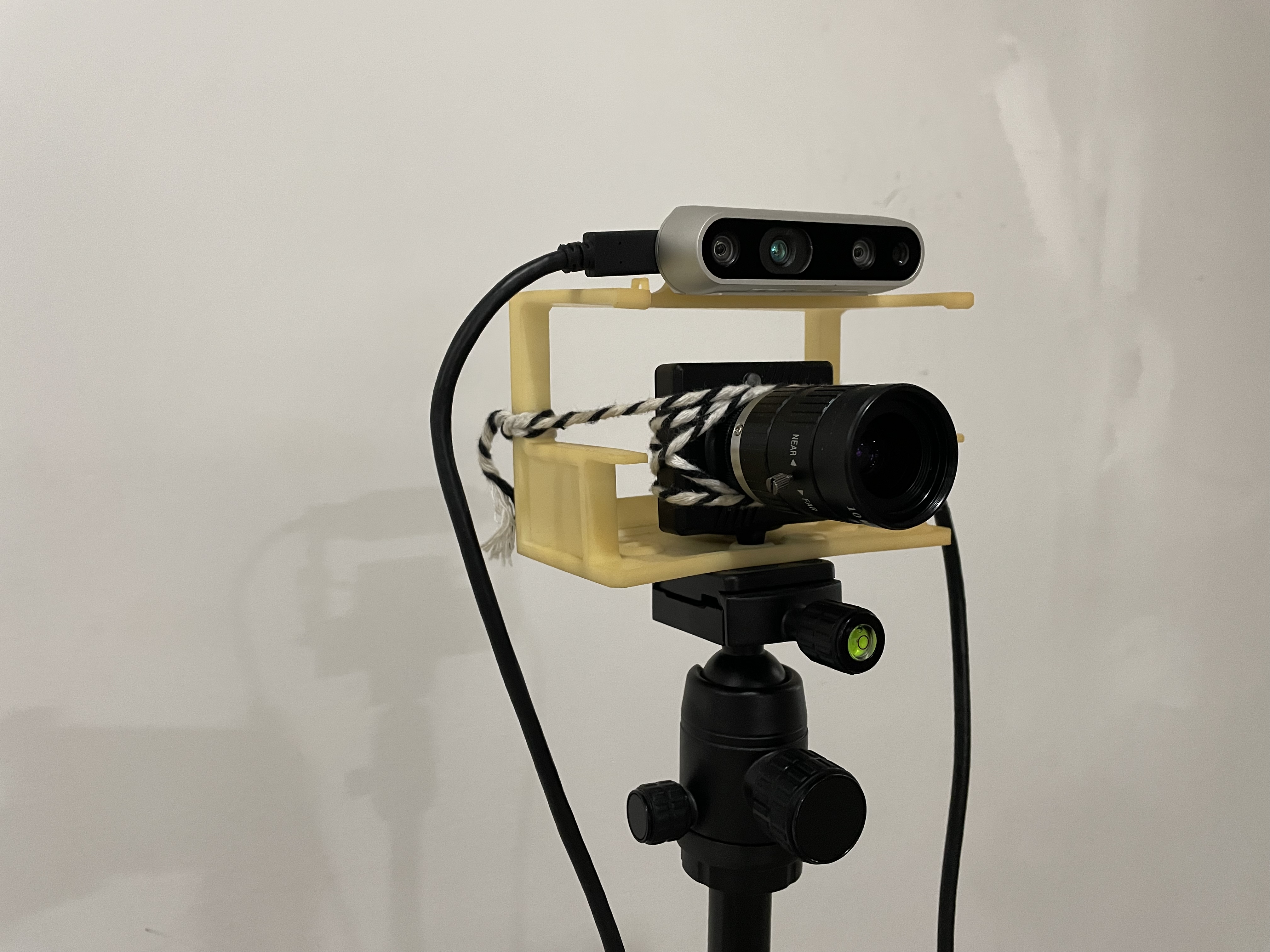}
\caption{Multiple camera system for data acquisition, with the RealSense D435i on top and CeleX-V below.}
\label{camera_pose}
\end{figure}

\begin{table}[t]
\caption{Lists of recorded human actions performed with low/medium/fast speeds \label{action_list}}
\centering
\begin{tabular}{c|l}
\hline
\hline
Speed  & Actions \\
\hline
slow   & \begin{tabular}[c]{@{}l@{}} walking, picking up, crawling, sweeping, mopping, \\ playing shuttlecock \end{tabular} \\
\hline
medium & \begin{tabular}[c]{@{}l@{}} squat jumping, frog jumping, boxing, cartwheel, \\ rope skipping, supine jumping, kicking, ball throwing, \\ spinning, throwing \end{tabular} \\
\hline
fast   & \begin{tabular}[c]{@{}l@{}} running, bungee jumping, open and closed jumping, \\ crotch high five, cycling, alternating squat jump, \\ spreading arm big jump, assisted long jump, Bobbi jump \end{tabular}\\
\hline
\hline
\end{tabular}
\end{table}

\textbf{Camera calibration}. In our multiple camera system, we use three imaging cameras, including event camera, depth, and RGB cameras, where depth and RGB cameras are integrated into one single RGB-D camera D435i. We need to calibrate the intrinsic and extrinsic parameters of these cameras in order to obtain the transformations between camera spaces. First, we calibrate the intrinsic parameters of the event camera, where particularly gray-scale images rather than event streams are used during the calibration. For the RGB-D camera, we directly use the intrinsic and extrinsic parameters provided by the manufacture for RGB and depth cameras. Lastly, we calibrate the relative mapping matrix of the event camera space to the RGB-D camera space while the intrinsic and extrinsic parameters of RGB-D cameras are fixed during calibration. In this way, we can directly project the joint locations in the 3D space of RGB-D camera to the 2D space of event camera.





\textbf{Annotation}. Currently, there are few event-based datasets for human pose estimation. One main reason is that it is not easy to annotate on event streams when dataset is captured outdoors. Most of the existing datasets are collected indoors using a set of motion capture sensors to assist annotation, which limits the diversity of human actions. Therefore, we use a simple but effective multiple camera system that are easily to annotate the joint locations. In our multiple camera system, we need to calibrate the intrinsic and extrinsic parameters of these cameras in order to obtain the transformations between camera spaces. With the calibration, we first manually annotate the ground truth of 2D joints on the RGB frames. The depth of each 2D joint is then obtained by warping its corresponding depth image to the RGB frame. Accordingly, we can have 3D joint locations in the 3D space of RGB-D camera by using the intrinsic parameters of depth camera. Lastly, it is straightforward to obtain the 2D joint locations in event camera space by projecting the annotations in RGB-D camera space based on the calibrated matrix. It should be noted that since that we integrate the event stream to a sequence of successive frames in a fixed time interval, we temporally align the RGB-based annotations to the event frames according to their frame rates as in \cite{calabrese2019dhp19}. To implement the annotation, we developed a interactive GUI tool as illustrated in a figure of Appendix. In addition, some exemplar annotated action samples can be found in the Appendix.

\textbf{Data comparison}. We compare our dataset with two other existing datasets. As shown in \tablename~\ref{compare_dataset}, although our data is not the largest one in terms of the number of frames, we have a more diversity in action classes, as listed in 
\tablename~\ref{action_list}. Moreover, our dataset is acquired in multiple outdoor scenes under varying environments rather than a fixed environment. Our dataset also can offer the possibility of 3D pose estimation in future, because we have the annotations of 3D joint locations by using the depth camera.

\textbf{Event frame and video generation}. In order to leverage the frame-based deep learning techniques, we need to convert asynchronous events into frames, referred as event frames. To do that, we follow the strategy in \cite{Maqueda2018} to simply accumulate the events at the fixed time interval of 8.333 $ms$ in a pixel-wise manner to obtain 2D histograms of events as an event frame. According to this procedure, around 82K event frames are generated in ours CDEHP dataset. A set of consecutive frames generated from each event stream compose an event video. Using these videos, we segment them into a number of successive short video clips of the length $T$ same with the temporal length of our recurrent network. These short video clips are then used to train and test our model. 

\begin{table}[t]
\caption{Existing event-based human pose datasets are compared in terms of the number of subjects (Sub\#), the number of actions per subject (Act\#), the number of frames (Frame\#), and multi-modality (MM). The shooting scenes are also listed to compare.\label{compare_dataset}}
\centering
\begin{tabular}{c|c|c|c|c|c}
\hline
\hline
Dataset & Sub\# & Act\# & MM & Frame\# & Scenes \\
\hline
DHP19 \cite{calabrese2019dhp19} & 17 & 33 & No & 87k & indoor \\
\hline
MMHPSD \cite{zou2021eventhpe} & 15 & 12 & Yes & 240k & indoor \\
\hline
CDEHP(ours) & 20 & 25 & Yes & 82k & outdoor \\
\hline
\hline
\end{tabular}
\end{table}  


\section{Experiments}
\label{sec:experiments}
In this section, we first introduce our experimental setup. Second, extensive experiments for ablation studies are carried out to analyze and verify the effectiveness of the proposed method. Finally, we compare our method to existing state-of-the-art methods on the DHP19 MMHPSD, and CDEHP dataset. The results demonstrate the effectiveness and strength of our method in event-based human pose estimation.

\subsection{Experimental Setup}
\label{parameter}
\textbf{Datasets.}
We evaluate our method on three datasets, DHP19\cite{calabrese2019dhp19}, MMHPSD\cite{zou2021eventhpe}, and CDEHP. DHP19 is the first dataset collected for event-based human pose estimation. It contains a total of 33 actions, where each action is performed 10 times by 17 subjects (12 female and 5 male). The DHP 19 dataset was recorded from four event cameras located in different four views. We only adopt the event data from a single camera view (camera view 3 as illustrated in \cite{calabrese2019dhp19}) to evaluate our approach, while our paper focuses on the 2D pose estimation from event data. Following the previous work \cite{calabrese2019dhp19}, we split 17 subjects into 12 subjects for training and 5 subjects for testing for evaluation. MMHPSD is a latest event-based human pose estimation dataset which was initially created for 3D shape estimation, where 4 different types of visual sensors, including one event camera, one polarization camera, and five RGB-D cameras, are used to capture the actions. We utilize 2D joints in the pre-processed data, issued publicly by the authors of \cite{zou2021eventhpe}, as the pose annotations, which are obtained by projecting the fine-tuned 3D shapes constructed from the annotations of multi-view RGB frames \cite{zou2021eventhpe}. In MMHPSD, there are 15 subjects performing 3 groups of actions (21 actions in total) for 4 times, where each group includes actions with slow/medium/fast speed respectively. The detailed action classes of the MMHPSD dataset can be found in \cite{zou2021eventhpe}. Following the work \cite{zou2021eventhpe}, we split 15 subjects into 12 subjects for training and 3 subjects for testing. For our CDEHP dataset, its details has been introduced in Section~\ref{sec:cedhp}. For evaluation, we split it a training set from 15 subjects and a testing set from the remained 5 subjects.

\textbf{Implementation.} 
In the experiments, all event frames are cropped to a fixed size of $256 \times 256$ with the human bodies set at center. Random rotation is simply used for the data augmentation. During training, the temporal length of our recurrent network is set to be 16 (i.e., $T=16$), which is found to be large enough to obtain sufficient information for completing lost events, achieving the best performance. Adam optimizer with the learning rate of $5e$-5 and the weight decay of $1e$-4 is used to optimize the learning process. If the training loss does not continue to decrease in $5$ epochs, the learning rate is then reduced by half until the learning rate is less than $1e$-6 and then the training ends. The model is carried out on four NVIDIA 2080Ti GPUs. 

\textbf{Evaluation metrics.} 
The commonly used standard evaluation metric is the object keypoint similarity (OKS)  \cite{Cheng_2020_CVPR}, which is calculated based on the Euclidean distance between a detected joint and its ground truth. We report a set of standard average precision scores: $\text{AP}^{50}$ (AP at OKS = 0.50), $\text{AP}^{75}$ (AP at OKS = 0.75), and $\text{AP}$ (AP at $\text{OKS} = 0.50, 0.55, ..., 0.90, 0.95$). In addition, we also adopt the percentage of correct keypoints (PCK) metric, which reports the percentage of correct keypoint detection. A detected joint is considered as correct if it falls within a normalized distance to the ground truth. Nevertheless, on the DHP19 dataset, the PCK and AP tend to saturate on this dataset by using even a very simple network, since that they both can reach almost 100\% when every model listed in \tablename~\ref{ablation_compare} has just started the training for a few epochs. Therefore, we employ the mean per joint position error (MPJPE) as the evaluation metric, which represents the average Euclidean distance between ground truth and prediction, which is usually measured with pixel distances in 2D image space, i.e., $\frac{1}{K}\sum_{k}^{K}\left \| p_i-p^{*}_{i} \right \|$. $p_k$ and $p^{*}_{k}$ denote the ground truth and predicted position of the $k$-th joint in image space.

\begin{figure}[t]
\centering
\subfloat[]{\includegraphics[width=2in]{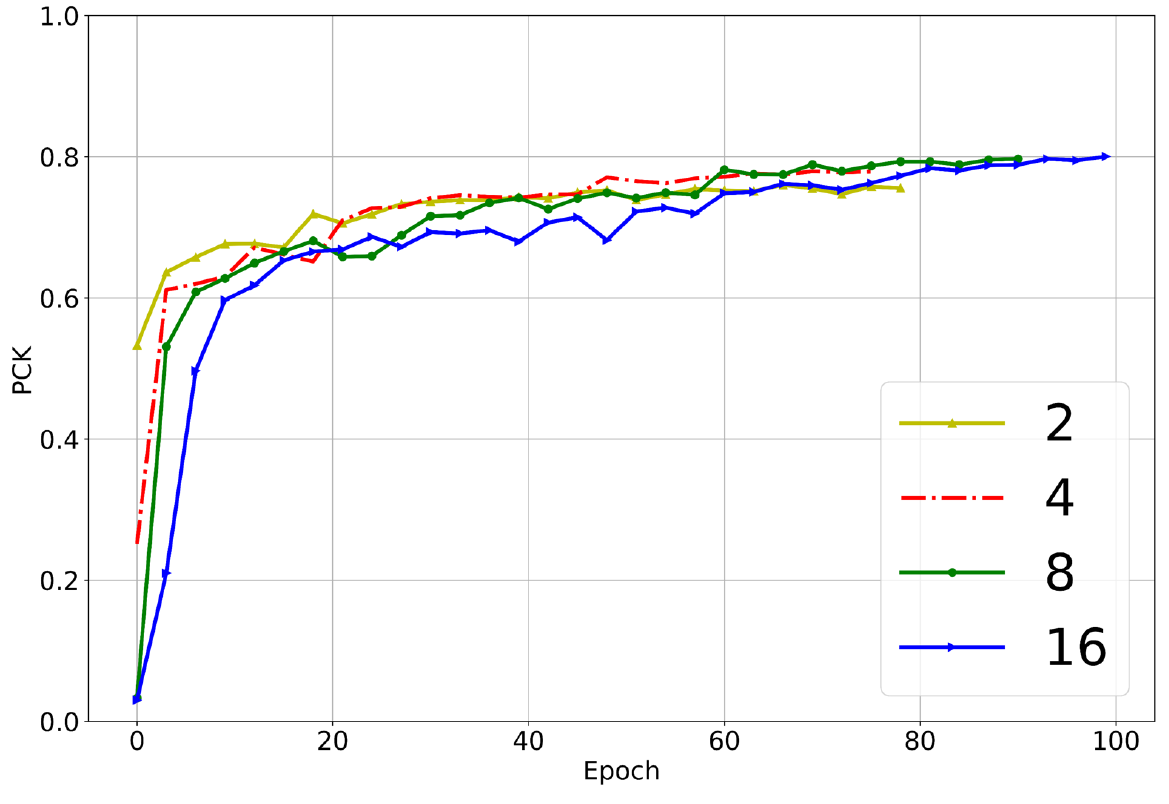}%
\label{fig4a}}
\vfil
\subfloat[]{\includegraphics[width=2in]{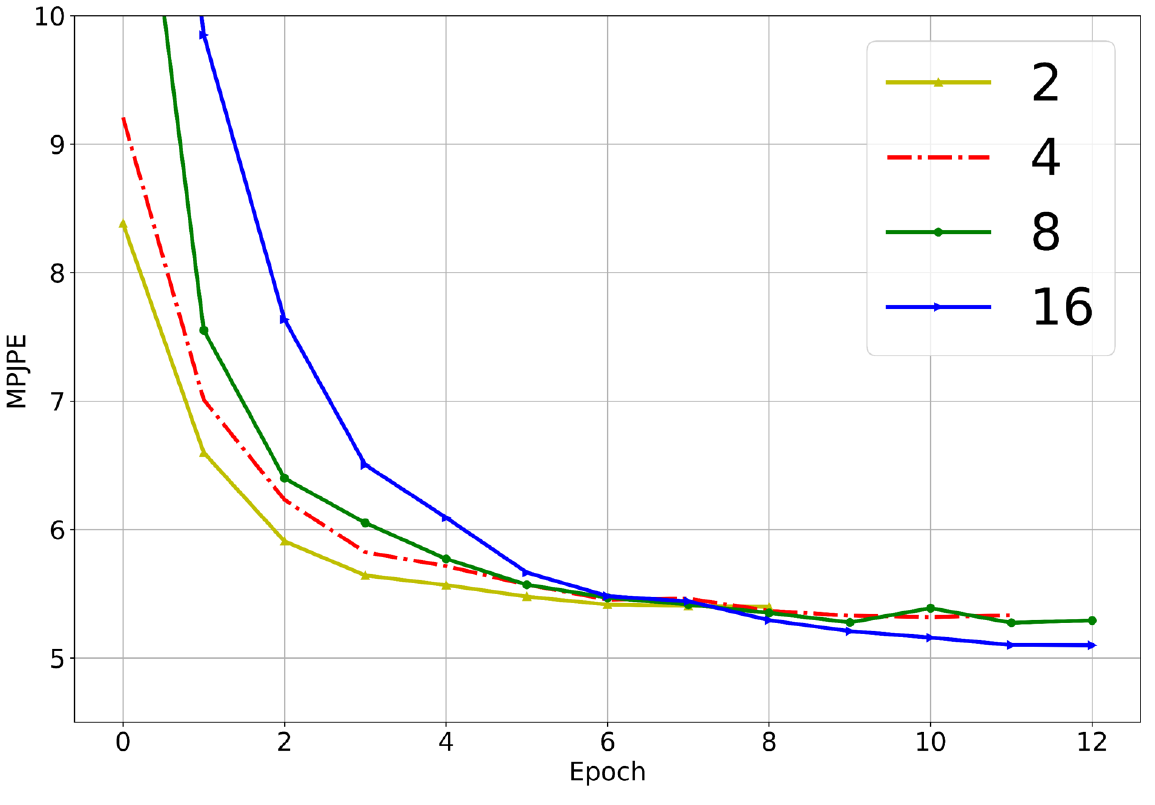}%
\label{fig4b}}
\vfil
\subfloat[]{\includegraphics[width=2in]{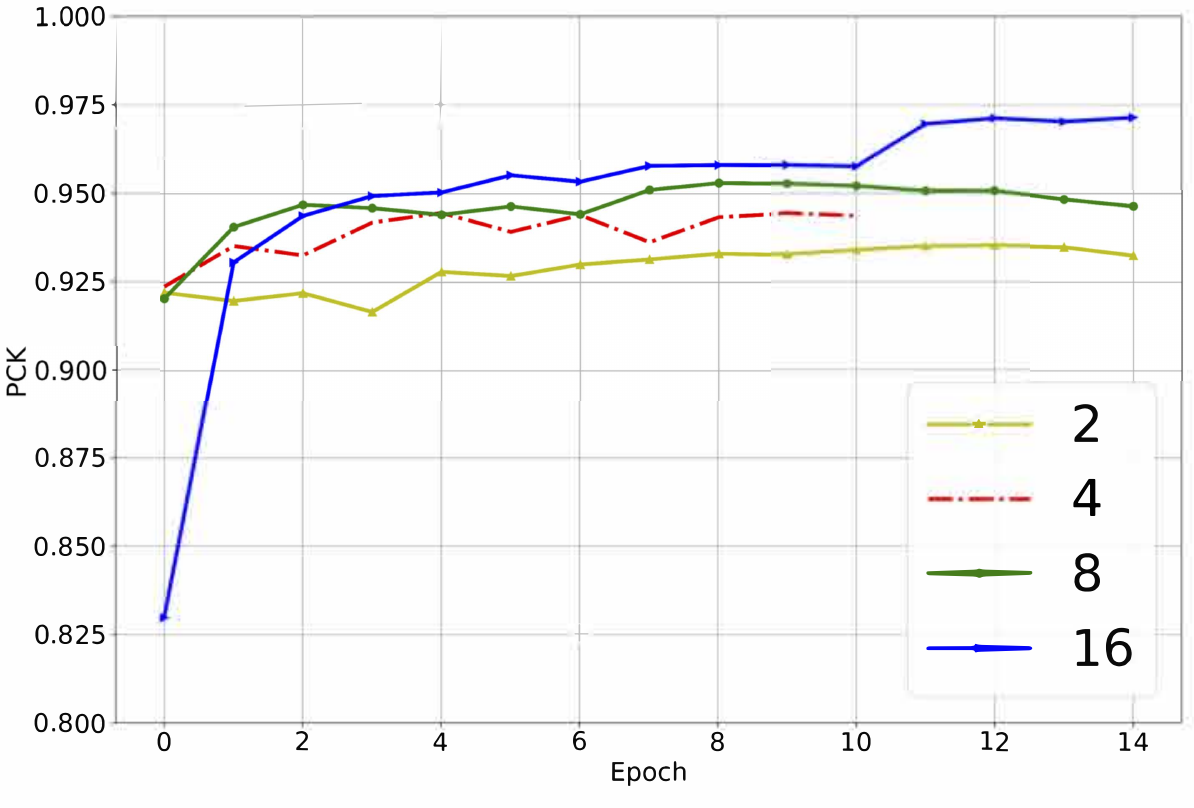}%
\label{fig4c}}
\caption{The model performance using different temporal lengths in our recurrent network on (a) CDEHP dataset, (b) DHP19 dataset, and (c) MMHPSD dataset, respectively. PCK metric is used on CDEHP and MMHPSD, while MPJPE is used on DHP19. Best viewed in color.}
\label{effects_of_length_fig}
\end{figure}

\subsection{Experimental Analysis}
\textbf{Effect of the temporal length.} Temporal lengths $T$ in Eq.~\ref{eq5} is a hyper-parameter that refers to the number of successive frames input to the recurrent network for the training. We experiment with the varying temporal lengths, i.e., $T=2, 4, 8, 16$, and report the results in terms of the PCK metric on the three datasets, as recorded in \figurename~\ref{effects_of_length_fig}. The figure shows the effects of different temporal lengths on the model performance with increasing epochs in the training process. We can observe the best performance on CDEHP dataset is achieved with $T=8,16$, while on other two datasets the best results are achieved with $T=16$. The results with the shorter temporal lengths decrease significantly especially when $T=2,4$. This study reveals that with the proposed densely connections in a long range are effective to pass previous useful information to help predict the pose at the current frame. But, it doesn't mean that the longer the temporal length is, the better performance is, while on CDEHP and DHP 19 datasets we can see the performance with $T=8$ is close to the one with $T=16$. It exhibits that the information of the preceding frames that are very far away from the current frame does not contain effective information as useful as the near frames. Meanwhile, a longer temporal length could bring a higher computational cost accordingly. Therefore, we do not try to set up a higher temporal length over the current setting $T=16$, as we set $T=16$ in this paper.

\textbf{Contribution of each component.} We conduct extensive experiments to study the effect of each component in our model. First, we insert a convolutional LSTM module between \textit{ConvNet1} and \textit{ConvNet2} to simply create a standard recurrent neural network for modeling the temporal dependency across frames, as stated in Eq.~\ref{eq2}. We consider this model as our baseline, denoted as RNN. Next, we incorporate one single connection between every two adjacent frames in the baseline model, where the output heatmap at the last frame are passed to the current frame to concatenate with the features as newly input to the LSTM module. We denote this model as tThinRNN. Instead of one single connection, the dense connection is introduced to the baseline model, as described in Eq.~\ref{eq3}, which is denoted as tDenseRNN-w/o-AT. Lastly, the attention mechanism is furthermore incorporated into the tDenseRNN-w/o-AT model to form our full model, as stated in Eq.~\ref{eq5}. We use tDenseRNN to denote our full model. For all above models, we fix the backbone of the encoder in \textit{ConvNet1} as ResNet-18.


\begin{table*}
\caption{Comparison with state-of-the-art methods on CDEHP, MMHPSD and DHP19 dataset. Best in bold, second-best underlined. The input size for all methods is $256\times256$\label{ablation_compare}}
\centering
\begin{tabular}{l|l|llll|llll|l}
\hline
\hline
 \multirow{2}{*}{Method} & \multirow{2}{*}{Backbone} &\multicolumn{4}{c|}{CDEHP} & \multicolumn{4}{c|}{MMHPSD \cite{zou2021eventhpe}} & DHP19 \cite{calabrese2019dhp19} \\ \cline{3-11} 
 &  & AP $\uparrow$ & $\text{AP}^{50}$ $\uparrow$ & $\text{AP}^{75}$ $\uparrow$ & PCK $\uparrow$ & AP $\uparrow$ & $\text{AP}^{50}$ $\uparrow$ & $\text{AP}^{75}$ $\uparrow$ & PCK $\uparrow$ & MPJPE $\downarrow$\\
\hline
Hourglass \cite{newell2016stacked}   & 8-Stack HG &   75.87 & 91.78 & 59.47 & 71.32 & 76.47 & 94.88 & 65.55 & 91.74 & 7.18 \\
SimpleBaseline \cite{xiao2018simple} & ResNet-18  &   77.51 & 93.10 & 63.20 & 73.60 & 77.16 & 95.12 & 67.73 & 91.84 & 7.15 \\
HigherHRNet \cite{Cheng_2020_CVPR}   & HRNet-W32  &   75.60 & 91.65 & 57.95 & 71.56 & 78.18 & 95.53 & 70.62 & 92.14 & 7.02 \\
LSTM-CPM \cite{luo2018lstm}          & CPM        &   59.37 & 67.63 & 28.10 & 49.07 & 40.99 & 39.28 & 3.66  & 54.75 & 7.36 \\
DKD \cite{nie2019dynamic}            & ResNet-18  &   78.97 & 95.37 & 67.36 & 76.79 & 81.07 & 97.44 & 77.90 & 94.41 & 5.40\\
DCPose \cite{Liu_2021_CVPR}          & ResNet-18  &   77.56 & 93.65 & 63.18 & 74.80 & 81.97 & 97.45 & 80.62 & 95.02 & 6.62 \\
FAMI-Pose \cite{liu2022temporal}     & HRNet-W32  &   78.40 & 93.23 & 67.43 & 76.90 & 80.31 & 96.53 & 76.45 & 93.54 & 5.53 \\
\hline
RNN                                  & ResNet-18 &   77.97 & 94.09 & 64.63 & 75.79 & 84.15 & 98.23 & 85.26 & 95.76 & 5.36 \\
tThinRNN                             & ResNet-18 &   78.18 & 94.82 & 64.53 & 76.00 & 84.90 & 98.65 & 87.48 & 96.15 & 5.32 \\
tDenseRNN-w/o-AT                     & ResNet-18 &   79.54 & 95.06 & 69.74 & 78.95 & \underline{86.82} & \underline{98.83} & \underline{90.55} & \underline{97.00} & 5.17 \\
tDenseRNN                            & ResNet-18 &   \underline{80.18} & \underline{95.51} & \underline{71.50} & \underline{79.70} & \bf{86.96} & \bf{99.09} & \bf{91.77} & \bf{97.14} & \underline{5.08} \\
tDenseRNN+PoseAug                       & ResNet-18  &   \bf{82.22} & \bf{96.60} & \bf{77.23} & \bf{82.66} & -- & -- & -- & -- & \bf{4.55} \\              
\hline
\hline
\end{tabular}
\end{table*}

From the results on all datasets, as listed in \tablename~\ref{ablation_compare}, we can see using RNN already serves as a simple and strong baseline, achieving 5.36 MPJPE on DHP19, 84.15 AP on MMHPSD, and 77.97 AP on CDEHP dataset. Especially on MMHPSD and DHP19 dataset, our all models can achieve a set of similar performance with marginal differences, because actions in MMHPSD and DHP19 are not very challenging so that our models are easy to saturate on these two datasets. Therefore, in the following experimental study, we only analyze the results on our CDEHP dataset. For tThinRNN, we can find that it does not improve much in performance and even its AP drops a little (-0.30 AP) in comparison with the RNN model. By adding dense connections, tDenseRNN-w/o-AT can outperform tThinRNN by a distinct margin of +1.63 AP. Furthermore, tDenseRNN with the attention mechanism achieve the best performance, where AP is improved from 79.30 to 80.24. In terms of PCK, we even improve the performance by a large margin of +1.30 PCK compared to tDenseRNN-w/o-AT. These results verify the strength of our dense connections with the attention mechanism for the pose estimation from event video.

\begin{figure}[bt]
\centering
\subfloat[]{\includegraphics[width=1.5in]{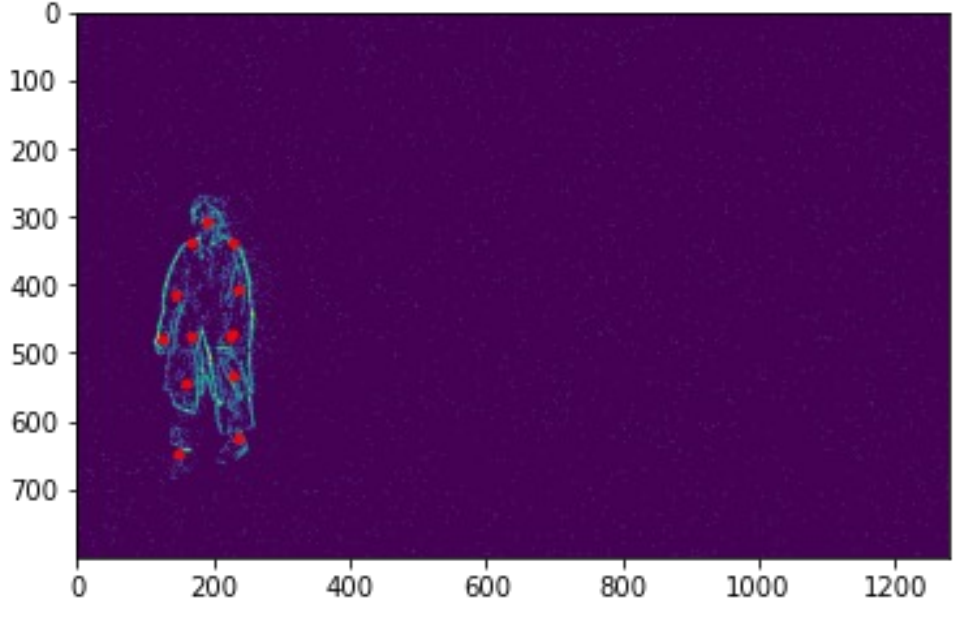}%
\label{fig5a-1}}
\hfil
\subfloat[]{\includegraphics[width=1.5in]{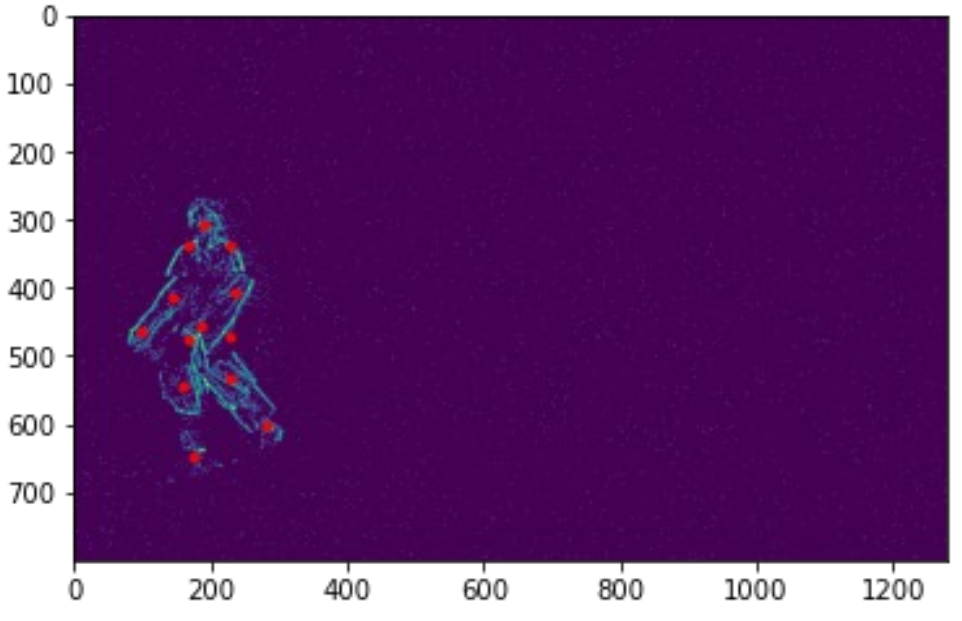}%
\label{fig5b-1}}
\caption{An example of the pose augmentation on pose events. (a) An original pose. (b) The augmented pose. Best viewed in color.}
\label{pose_aug_example}
\end{figure}

\textbf{Pose augmentation.} Motivated by \cite{gong2021poseaug}, we introduce a simple approach to augment the training poses of our dataset towards a greater diversity and thus improve the generalization of our model. Specifically, we augment the pose events $\{u_i,v_i\}$ in an event frame $I_t$ by rotating those events by the joints of knees and elbows, generating new pose events, as illustrated in \figurename~\ref{pose_aug_example},
\begin{align}
&(u_i^a,v_i^a) = (u_i-u_c,v_i-v_c)
\begin{pmatrix}
cos\theta   & sin\theta\\ 
 -sin\theta & cos\theta
\end{pmatrix}
+\left ( u_c,v_c \right )
\label{eq7}
\end{align}
where $(u_i^a,v_i^a)$ is the new locations of events in the augmented poses and $(u_c,v_c)$ is the center point of rotation. $\theta \in \left(-90,90\right)$ is set randomly. 

Our model trained with the additional augmented training poses, denoted as tDenseRNN+PoseAug, can achieve significant improvements over the model tDenseRNN on CDEHP and HDP19 dataset, which demonstrates the diversity of training data is quite useful to obtain a greater generalization of the model. It should be noted that we do not augment the data for MMHPSD dataset, because the raw data of MMHPSD dataset have not been available publicly so far.




\subsection{Comparisons with State-of-the-Arts}

\textbf{Results on DHP19}. \tablename~\ref{ablation_compare} summarizes the results on the DHP19 datasets, where we can see our baselines and the full model tDenseRNN achieve the best performance compared to other competitors. However, on DHP19 all methods achieve nearly same level of performance since that there are only pixel errors of $1\sim2$ in terms of the used metric of 2D MPJPE, while they can produce almost the result of 100\% in terms of PCK and AP metrics on DHP19. Therefore, obviously the DHP19 cannot be used to evaluate our method, because the included actions in this dataset are too simple so that it has not be a challenging dataset as the evaluation benchmark for pose estimation. 


\begin{table*}[tbh]
\begin{threeparttable}[t]
\caption{Action-wise result comparison on CDEHP dataset in terms of the AP metric. \textbf{Slow} actions, \textbf{Medium} actions, and \textbf{Fast} actions are included in the top, middle, and bottom parts, respectively, while we separate the table into three parts in terms of the action speed. Best in bold, second-best underlined. \label{action_wise}}
\centering
\begin{tabular}{c|l|l|l|l|l|l|l|l|l|l|l}
\hline\hline
 \multirow{3}{*}{\bf{Action}} &\multicolumn{10}{c}{\bf{Method}} \\ \cline{2-12} 
 &Hourglass  & \begin{tabular}[c]{@{}l@{}} Simple \\Baseline \end{tabular}  & \begin{tabular}[c]{@{}l@{}} Higher \\HRNet \end{tabular} & \begin{tabular}[c]{@{}l@{}} LSTM-\\CPM  \end{tabular} & DKD & DCPose & \begin{tabular}[c]{@{}l@{}} FAMI-\\Pose  \end{tabular} & RNN & tTR\tnote{1} & tDR-w/o-A\tnote{2} & tDR\tnote{3} \\
 \hline
walking  &68.99 &77.63  &73.40  &42.11  &\bf{80.64}  &77.50  &75.57 &77.04  &71.78  &77.12  &\underline{80.52}   \\ \cline{1-1}  
picking up &73.39   &73.94  &72.14  &56.86  &\underline{76.66}  &73.58  &75.36 &75.44 &72.53  &\bf{77.06}  &75.01   \\ \cline{1-1} 
crawling &62.79     &64.39  &62.65  &31.27  &66.79  &65.04  &61.43 &64.63   &66.98  &\underline{67.69}  &\bf{70.02}   \\ \cline{1-1} \  
sweeping &73.94      &75.99  &73.08  &61.92  &76.35  &73.87  &\bf{78.68} &75.57  &76.06  &77.18  &\underline{78.23} \\ \cline{1-1} \  
shuttlecock kicking & 86.25      &84.26  &84.88  &71.64  &86.69  &84.62  &\bf{89.08} &85.01   &86.27  &85.62  &\underline{87.54}   \\ \cline{1-1} \cline{1-12}
\it{Average}    &\it{73.07}  &\it{75.24}      &\it{73.23} &\it{52.76} &\it{\underline{77.43}}  &\it{74.92} &\it{76.02} &\it{75.54}  &\it{74.72}  &\it{76.94}  &\it{\textbf{78.27}} \\ 
\hline\hline
squat jump &89.42    &89.35  &88.71  &78.49  &89.75  &87.80  &88.67 &89.51 &\bf{90.55}  &89.64  &\underline{89.93}  \\ \cline{1-1} \  
frog jump &79.99     &79.11  &79.45  &57.50  &82.63  &80.89  &82.24 &81.42 &82.36  &\bf{83.59}  &\underline{83.09}  \\ \cline{1-1} \  
boxing &81.50        &80.64  &75.25  &71.45  &80.66  &81.31  &77.80 &82.08 &83.32  &\underline{83.54}  &\bf{83.58} \\ \cline{1-1} \  
cartwheel &57.49      &58.06  &56.80  &36.67  &60.20  &59.48 &\bf{64.98} &61.27 &60.73 &61.72  &\underline{63.37}  \\ \cline{1-1} \  
rope skipping &75.30 &75.18  &73.10  &65.75  &76.76  &77.62  &76.05 &76.27 &75.90  &\underline{77.74}  &\bf{78.18}  \\ \cline{1-1} \  
sit-up jump &74.94   &74.75  &73.02  &59.31  &75.46  &73.22  &73.89 &74.53 &\underline{76.84}  &76.78  &\bf{77.86} \\ \cline{1-1} \  
kicking &69.45       &72.45  &72.48  &60.81  &75.99  &73.85  &75.47 &75.87 &\underline{76.14}  &75.66  &\bf{77.33} \\ \cline{1-1} \  
jump shot &74.30     &75.68  &72.59  &58.02  &77.90  &76.41  &78.85 &76.76 &79.30  &\bf{80.33}  &\underline{79.52}  \\ \cline{1-1} \  
spinning &69.11      &\underline{75.39}  &72.27  &56.00  &74.37  &73.54 &75.20 &73.15 &74.25  &\bf{75.98}  &74.26   \\ \cline{1-1} \  
throwing &74.20      &75.08  &70.94  &57.46  &74.49  &75.29  &75.54 &72.45 &77.14  &\bf{78.35}  &\underline{78.04}   \\ 
\cline{1-1} \cline{1-12}
\it{Average} &\it{74.57}      &\it{75.57}  &\it{73.46}  &\it{60.15}  &\it{76.82}  &\it{75.94}   &\it{76.87} &\it{76.33} &\it{77.65}  &\underline{\it{78.33}}  &\it{\textbf{78.52}}\\ 
\hline\hline

jumping jack  &\bf{96.30}    &96.02  &95.54  &81.88  &95.88  &95.60 &95.53 &95.95  &95.55 &\underline{96.27}  &95.94  \\ \cline{1-1} \  
running  &73.90      &76.91  &73.60  &54.61  &79.50  &78.92  &78.89 &78.46  &80.05  &\underline{80.29}  &\bf{80.60} \\ \cline{1-1} \ 
burpee &72.81 &73.62  &71.45  &46.70  &75.38  &74.26 &73.50 &73.83 &76.32  &\underline{76.71}  &\bf{77.60} \\ \cline{1-1} \ 
mopping &69.19       &73.86  &70.15  &49.79  &74.05  &72.13 &70.75 &74.56  &72.97  &\underline{75.19}  &\bf{76.37}  \\ \cline{1-1} \  
cycling &72.31        &77.69  &77.12  &66.37  &\underline{79.93}  &75.95  &79.79 &79.14  &77.82  &79.56  &\bf{81.03} \\ \cline{1-1} \ 
big jump &92.11       &92.18  &92.66  &77.56  &93.01  &91.76 &\underline{93.29} &92.67  &92.98  &\bf{93.44}  &93.16 \\ \cline{1-1} \  
long jump &69.71     &70.13  &69.17  &54.03  &71.39  &72.23  &\bf{74.30} &70.77  &70.09  &72.78  &\underline{73.52}  \\ \cline{1-1} \ 
crotch high five  &87.89    &88.68  &87.79  &74.58  &89.07  &\underline{89.80}  &89.62 &87.83  &87.10  &88.77  &\bf{90.32}  \\ \cline{1-1} \  
alternate jumping lunge &77.88 &80.24  &77.28  &66.34  &80.60  &79.94  &78.98 &80.18  &81.50  &\bf{82.04}  &\underline{81.68}  \\ \cline{1-1} \  
jump fwd/bwd/left/right  &87.56  &87.87  &86.33  &80.19  &87.84  &87.58 &\bf{88.55} &87.00  &87.23  &\underline{88.32}  &87.37 \\ \cline{1-1} \cline{1-12}
\it{Average} & \it{79.97}        &\it{81.72}  &\it{80.11}  &\it{65.20}  &\it{82.66}  &\it{81.82}  &\it{82.32} &\it{82.04} &\it{82.16}  &\underline{\it{83.34}}  &\it{\textbf{83.76}}  \\ 
\hline\hline
\end{tabular}
\begin{tablenotes}
\item[1] tTR indicates tThinRNN, \item[2] tDR-w/o-A indicates tDenseRNN-w/o-AT, \item[3] tDR indicates tDenseRNN.
\end{tablenotes}
\end{threeparttable}
\end{table*}

\textbf{Results on CDEHP}. \tablename~\ref{ablation_compare} shows the comparisons of our model with state-of-the-arts on the CDEHP dataset, in terms of PCK and AP evaluation metrics. We can see that our tDenseRNN model achieves the best performance. Among those competitors, HigherHRNet \cite{Cheng_2020_CVPR} ,Hourglass \cite{newell2016stacked} and SimpleBaseline \cite{xiao2018simple} are three representative methods in image-based pose estimation, which achieve promising results of 75.60 AP, 75.87 AP and and 77.51 AP, respectively. DKD \cite{nie2019dynamic}, DCPose\cite{Liu_2021_CVPR}, and FAMI-Pose \cite{liu2022temporal} are three pioneering methods for video-based pose estimation, and they outperform those image-based methods by noticeable margins (+5.23 PCK, +3.24 PCK, and 5.34 PCK compared to HigherHRNet, respectively). This result implies that the temporal information can provide useful motion clues for event-based pose estimation. LSTM-CPM \cite{luo2018lstm} is also one of the strong baselines for video-based pose estimation method, which however does not produce an expected promising result (40.99 AP). This is may because it inherits the convolutional architecture from CPM \cite{wei2016convolutional} using a multi-stage refining scheme, which focuses on designing large receptive fields in multi-stage CNNs to capture long-range spatial dependencies that are not applicable for event frames with the absence of some body parts. Similar to DKD, DCPose and FAMI-Pose, RNN and tThinRNN introduce the recurrent module to model the temporal dependency achieving comparable results with them. By comparing them with our models tDenseRNN-w/o-AT and tDenseRNN, we can find our models always maintain high performance across all evaluation metrics, especially $\text{AP}^{75}$, thanks to the dense connections that are not considered in DKD, DCPose and FAMI-Pose, which only leverage a couple of adjacent frames.

\textbf{Results on MMHPSD}. As shown in \tablename~\ref{ablation_compare}, we have the similar observation with the results on CDEHP. video-based methods including ours can achieve promising results thanks to the temporal information considered, in comparison with image-based methods \cite{newell2016stacked,xiao2018simple,Cheng_2020_CVPR}. Different to the results on CDEHP, the overall better performance results can be obtained on MMHPSD across all methods, because the actions are simpler than CDEHP dataset and captured in a fixed environment. Meanwhile, our methods outperform the competitive methods including DKD, DCpose and FAMI-Pose \cite{nie2019dynamic,Liu_2021_CVPR,liu2022temporal} by a very noticeable margin, showing the strength of the temporal dense connection across frames.

\begin{figure*}[hbt]
\centering
\subfloat[]{\includegraphics[height=1.8in]{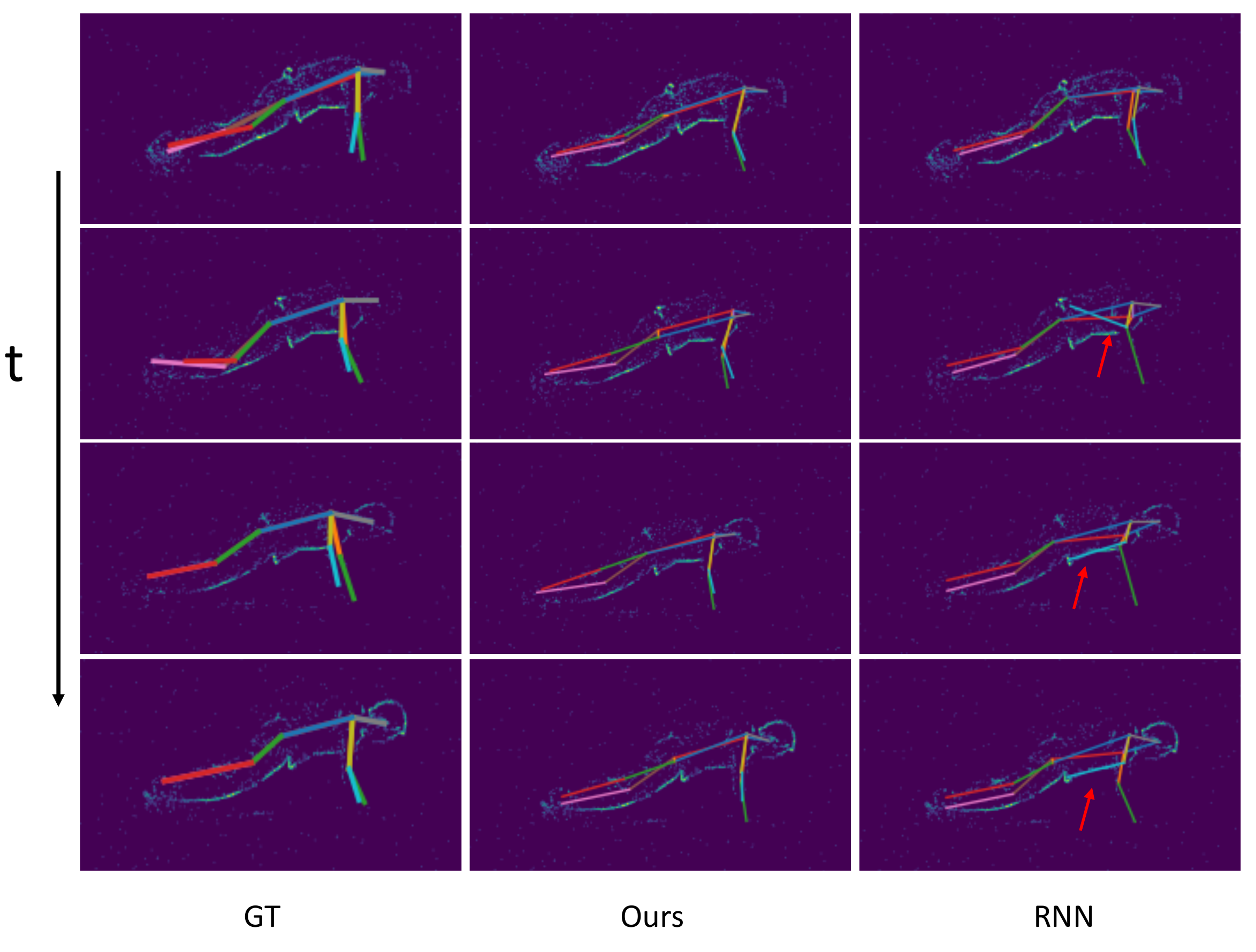}%
\label{fig5a}}
\subfloat[]{\includegraphics[height=1.8in]{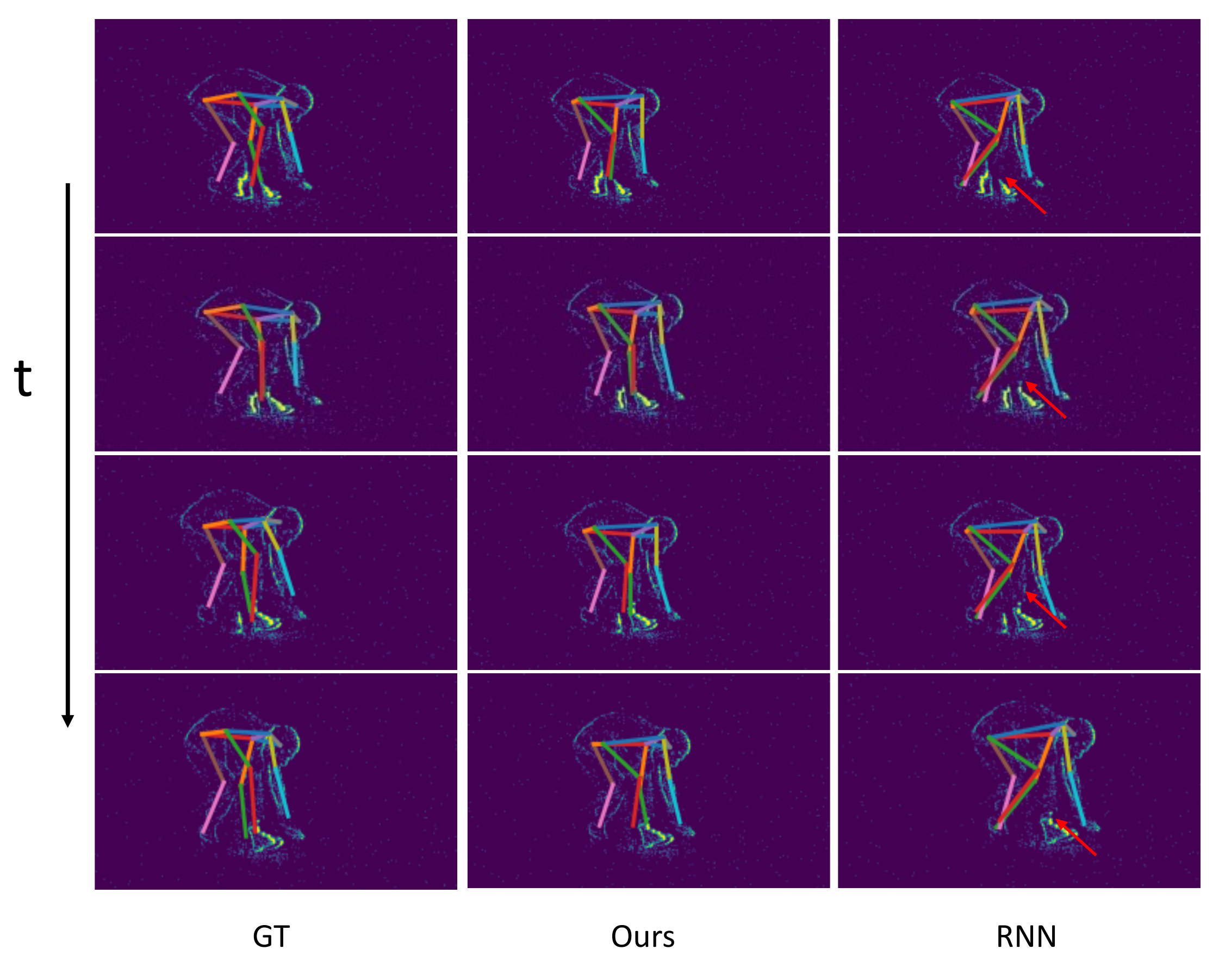}%
\label{fig5b}}
\subfloat[]{\includegraphics[height=1.8in]{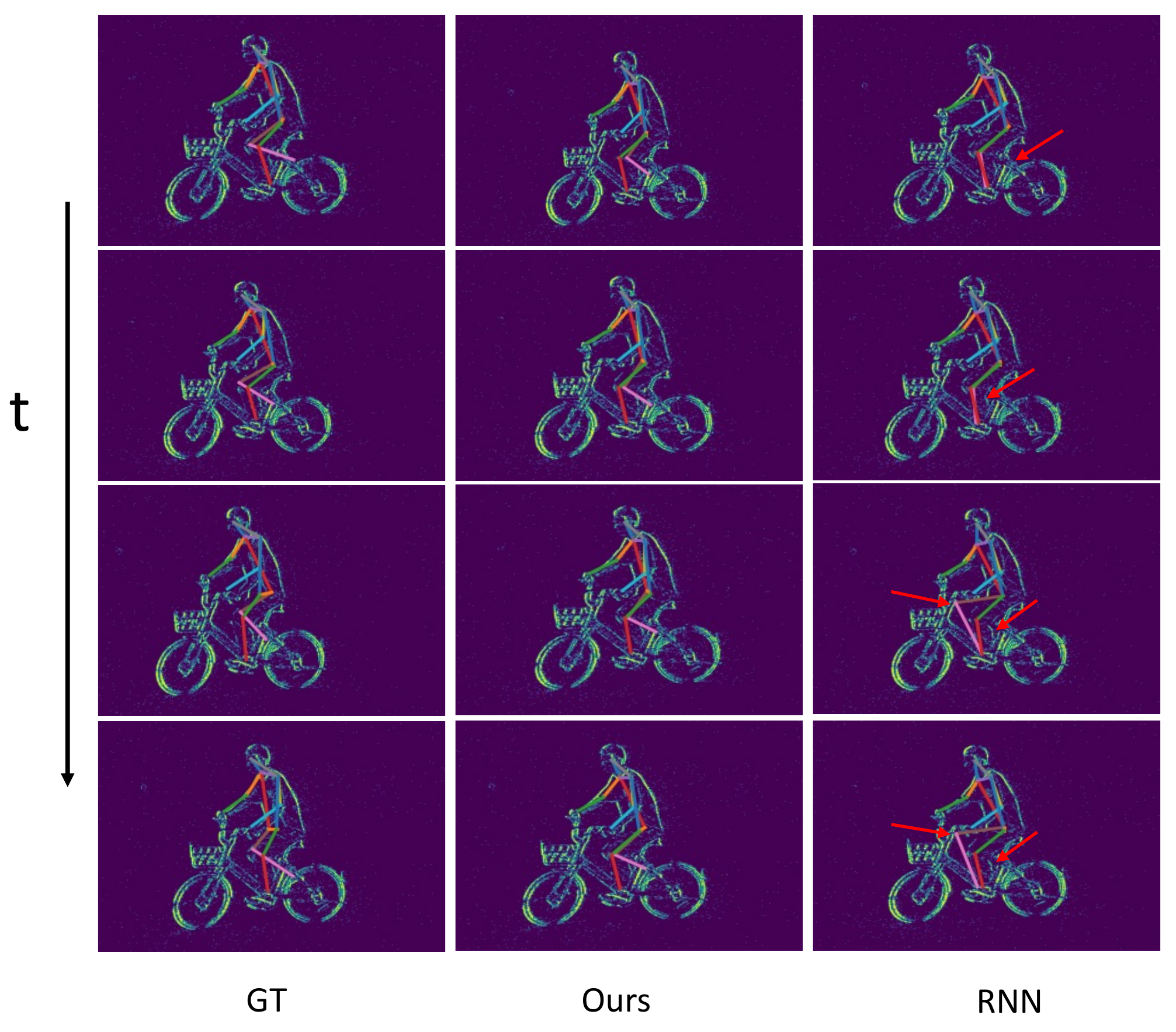}%
\label{fig5c}}
\hfil
\subfloat[]{\includegraphics[height=1.8in]{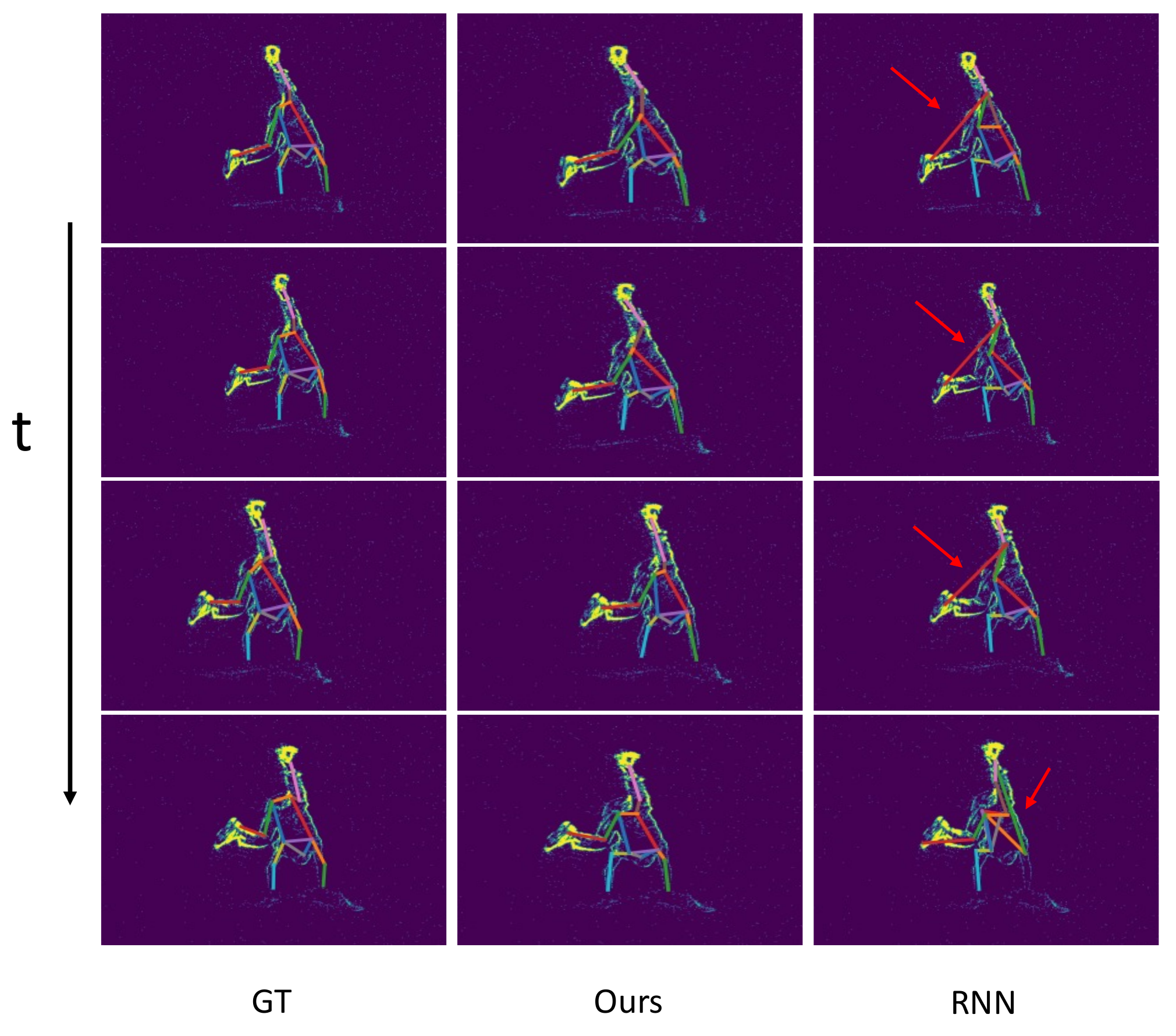}%
\label{fig5d}}
\subfloat[]{\includegraphics[height=1.8in]{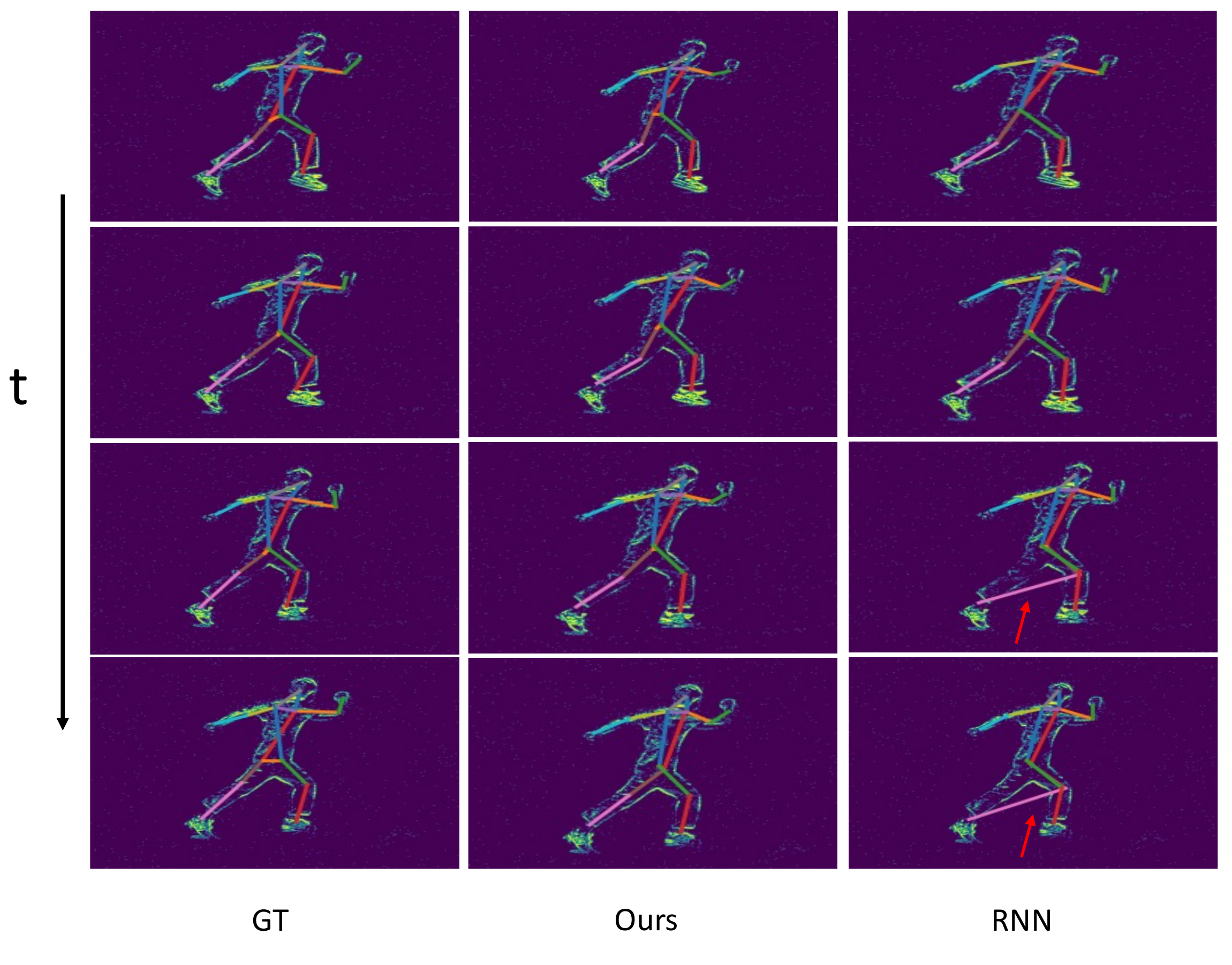}%
\label{fig5e}}
\caption{Qualitative results on the CDEHP dataset using our model tDenseRNN and the baseline model RNN, respectively. The samples are from actions of \textit{Burpee}, \textit{picking up}, \textit{cycling}, \textit{cartwheel}, and \textit{alternate jumping lunge}, respectively. Best viewed in color and $3\times$ zoom.}
\label{fig5}
\end{figure*}

\textbf{Action-wise result comparison on CDEHP dataset.} To analyze the results on various types of actions, we report the action-wise results on CDEHP dataset, as list in \tablename~\ref{action_wise}. We can observe that most of the challenges come from those slow/medium actions and complex actions, since that the former actions could produce more sparse event signals and the latter actions can bring severe self-occlusion of body parts in some cases. Therefore, a better performance can be obtained on those simple actions with fast speeds, as shown in \tablename~\ref{action_wise}. From the results, we can see that DKD \cite{nie2019dynamic} and FAMI-Pose \cite{liu2022temporal} are two methods that achieve promising results compared with our methods, while they can achieve the second/third best and even the best in some cases. In addition, almost all the methods can obtain the similar good performance on those simple actions, for example, \textit{squat jump}, \textit{jumping jack} and \textit{big jump}, while our methods (tDenseRNN) can particularly outperform other competitive methods by a large margin on those complex and slow actions, such as \textit{crawling, sit up jumping, burpee}, and \textit{cartwheel}. Moreover, for those actions that a few body part is moving, our method can also achieve superior results with a large performance improvement over other methods, such as \textit{throwing and mopping}. Moreover, we can obverse the performance improvements between our two baselines tDenseRNN and tThinRNN on those slow actions specifically, where the performance of TDenseRNN can usually gain noticeable improvements over tThinRNN thanks to the dense connections, such as \textit{walking, crawling, sweeping}. Overall, with our method the best or second results are obtained in all types of actions. This action-wise result demonstrates that our method, to some extent, can effectively alleviate the problem of the sparse and incomplete observations from event cameras. Meanwhile, it is observed that pose estimation from complex actions can also be addressed by our method.

\subsection{Visualization}
\textbf{Result Visualization.} Some qualitative results are shown in \figurename~\ref{fig5} to visualize the effectiveness of our model for event-based human pose estimation on CDEHP dataset compared to the baseline model RNN. We can observe that tDenseRNN can still accurately estimate the body joints when there are invisible body parts in event frames, as shown in \figurename~~\ref{fig5} (a)(d). In addition, for those complex actions, tDenseRNN can effectively handle the self-occlusions by encouraging the geometric consistency in the presence of fast and large-degree motion variations, as shown in \figurename~\ref{fig5} (b)(c)(e). These visualized results further verify the strength of our model tDenseRNN. 

\begin{figure}[hbt]
\centering
\subfloat[]{\includegraphics[height=2.5in]{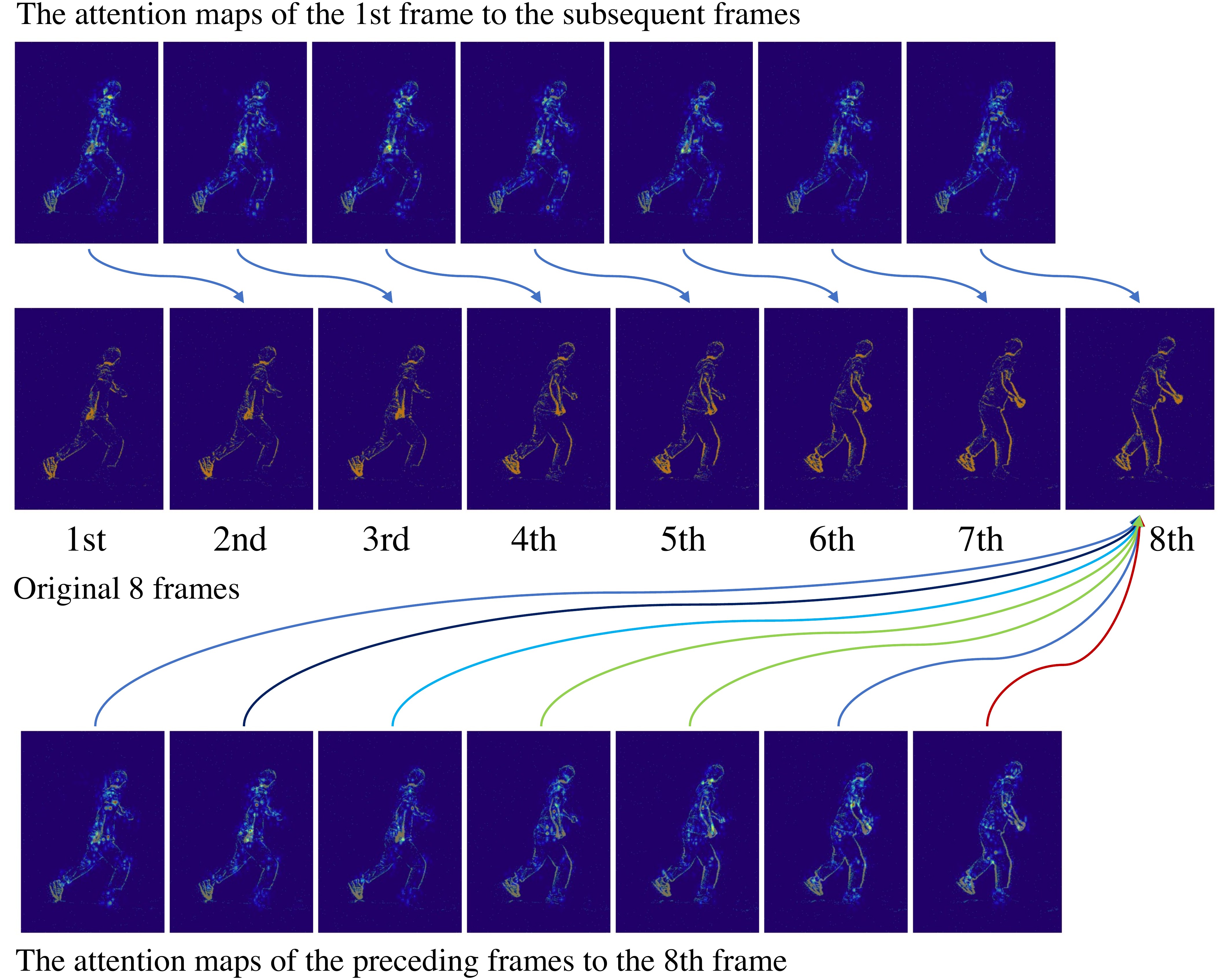}%
\label{fig8a}}
\hfil
\subfloat[]{\includegraphics[height=2.5in]{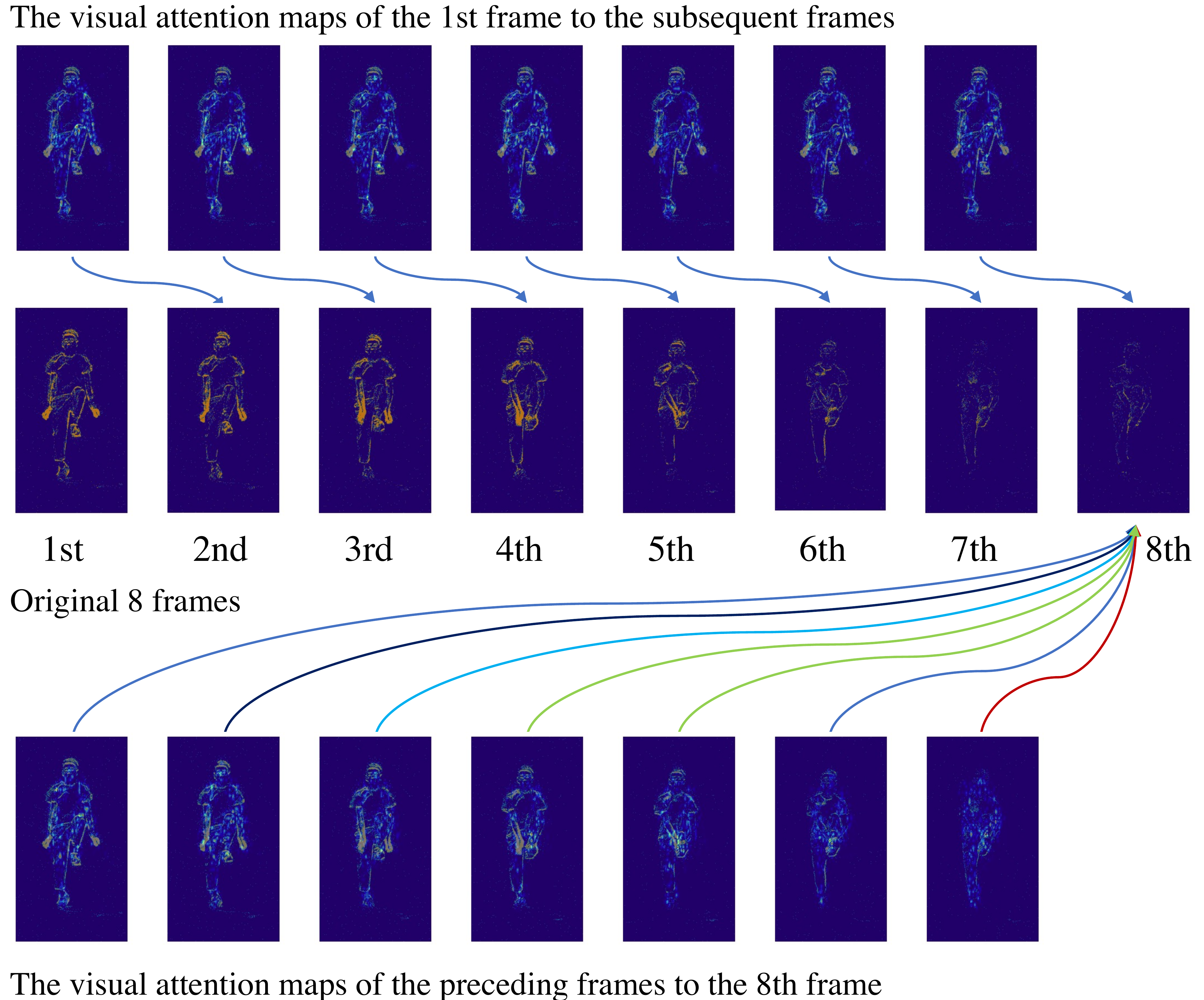}%
\label{fig8b}}
\caption{Attention Visualization. We take two short video clips (a) and (b) from action \textit{alternate jumping lunge} and \textit{Crotch high five} as two example sequences with 8 frames, respectively. In each visualized example, the top and bottom rows exhibit the visual attention regions on the corresponding original event frames, while the middle row shows the original event frames. The top shows the attention maps of the first frame to its subsequent seven frame. The bottom row shows the attention maps of the preceding seventh frames to the eighth frame. Best viewed in color and $3\times$ zoom.}
\label{attention_maps}
\end{figure}

\textbf{Attention Visualization.} We also visualize the attention maps on two example sequences to understand the key frames where information is densely propagated, as illustrated in \figurename~\ref{attention_maps}. In the first example as shown in \figurename~\ref{fig8a}, one of the foots are not complete or even disappeared at some frames. On one hand, the attention regions on the first frame are mainly distributed in those key joints, where especially the information of the foot joint can be propagated to help complete the foot information in the subsequent frames. On the other hand, each frame can receive attentions from a set of preceding frames, as shown in the bottom row of \figurename~\ref{fig8a}, where all the 7 preceding frames are combined to predict the human poses at 8th frame, since that body moving is continuous and temporally dependent. In the second example as shown in \figurename~\ref{fig8b}, the event signals at the 7th and 9th frames are very sparse. In this case, the preceding frames are very useful to alleviate the problem of incomplete event information for the human body. Especially, at the 1st frame one of the feet is visible, whereas at 5th and 6th frames that foot is not visible. Therefore, the attention regions at that foot area of the 1st frame to the 5th and 6th frames are much brighter to pass more information of that foot to the 5th and 6th frames. The visualization results show the effectiveness of the attention mechanism used in temporally dense connections.

{\subsection{Concluding Remarks}
Through extensive experiments above, we demonstrate the effectiveness of our method in addressing the challenges of event-based human pose estimation. As our approach converts event streams into a set of event frames, approaches for normal video-based human pose estimation, such as DKD, DCPose, and FAMI-Pose \cite{nie2019dynamic,Liu_2021_CVPR,liu2022temporal}, are directly employed to event frames as comparisons, achieving competitive results as shown in \tablename~\ref{ablation_compare} and \tablename~\ref{action_wise}. Even that, we do not adopt those complex schemes used in the video-based approaches to fuse and align neighboring frames with the current frame. Instead, we build a simple densely connected RNN architecture to achieve the best performance. Hence, we have demonstrated that leveraging a longer range of neighboring frames is a simple yet effective way to alleviate the problem of incomplete information in event-based human pose estimation. Furthermore, taking future frames into count, such done in DCPose and FAMI-Pose, is not a natural way to perform online tasks, since one of the advantages of using event cameras is their ability to operate in real-time manner.

\subsection{Limitations and future work}
This paper presents a temporal densely connected recurrent network for human pose estimation from event data. There are some potential limitations in our method. First, our method cannot work directly on raw event signals, which have to be converted into frames first. Second, for those unnormal actions like \textit{cartwheel}, where the human body is upside down rather than usually upright, we did not collect enough samples as many as the normal actions to train the model on CDEHP dataset. For future work, we will focus on estimating human poses from raw event signals by leveraging point-based methods, and try to extend the action samples to make the dataset more diverse. Finally, as shown in the results, and to the best of our knowledge, CDEHP is still the most challenging dataset in human pose estimation from event data. There is still plenty of room for improvement and problems remained, even though our method can help alleviate the problem brought by event signals in event-based human pose estimation, as the event signals are too sparse and irregular in spatio-temporal space when we try to encode complex actions. In the future, we will try to distill knowledge from other modalities, such as RGB and depth cameras, to help extract and encode event signals.
\section{Conclusion}
\label{sec:conclusion}

This paper presents a temporal densely connected recurrent network (tDenseRNN) for human pose estimation from event data. Particularly, it introduces a set of dense connections into the recurrent-based model to explicitly model both the sequential and non-sequential temporal dependency and geometric consistency among a set of consecutive event frames to address the problem of the absence of body parts during event streams. By such densely connected RNN, the adoption of even a very simple encoder-decoder CNN with a LSTM module can achieve superior performance to other state-of-the-art methods. tDenseRNN also introduces a spatio-temporal attention mechanism into the dense connections to weight the contributions of the preceding frames and the importance of joints within each preceding frame on the prediction at current frame, thus largely improving the performance. Moreover, to effectively evaluate our model, we collect a new large-scale event-based human pose dataset, which is captured from more challenging scenarios than the existing dataset that has been saturated by using our method. Experiments on three datasets demonstrate the strength and effectiveness of our method in addressing the event-based pose estimation, achieving the best performance. The dataset will be publicly online available soon for further research.

\bibliographystyle{IEEEtran}
\bibliography{egbib}

\newpage
\clearpage

{\appendix[Visualization of the CDEHP Dataset]

\begin{figure*}[b]
\centering
\includegraphics[width=7in]{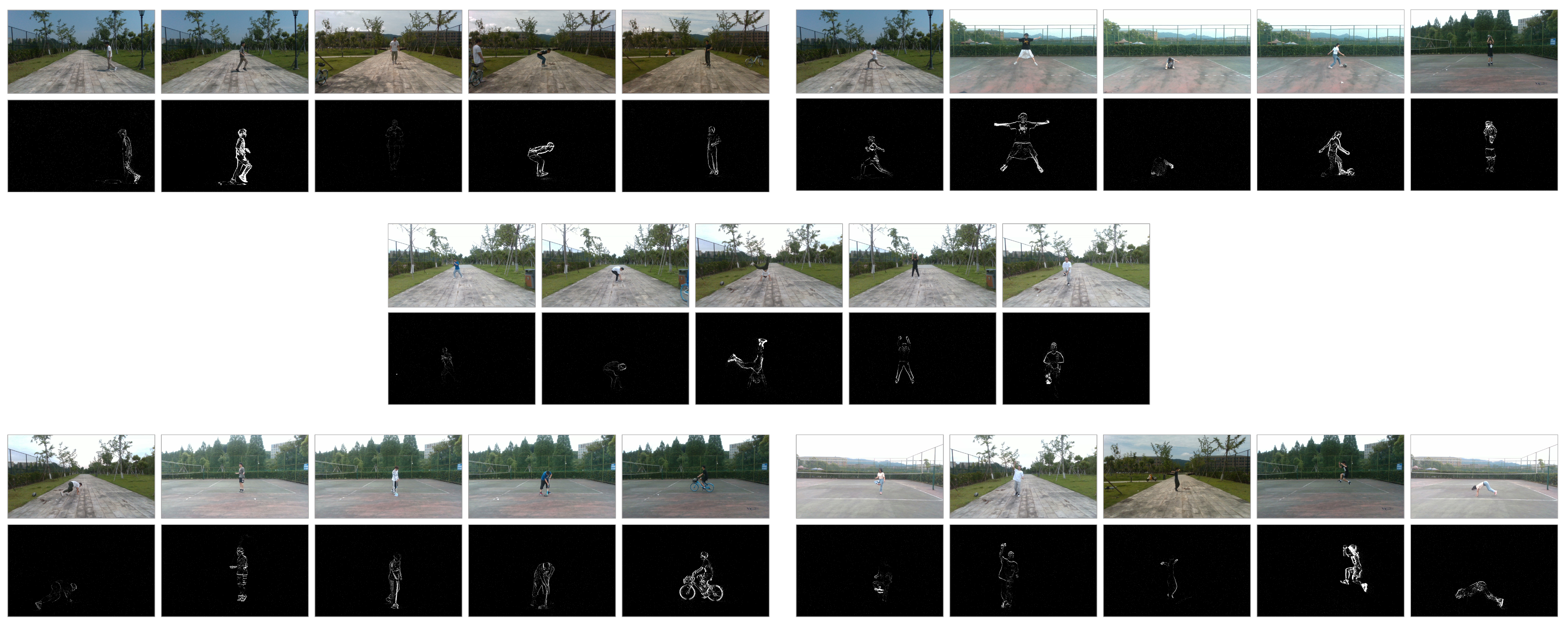}
\caption{Visualized samples for all action classes in CDEHP dataset. Best viewed by zooming in.}
\label{dataset_vis_all}
\end{figure*}

\figurename~\ref{dataset_vis} visualizes several action examples from RGB, depth, event modalities respectively. All actions are collected in various outdoor environments. As shown in the figure, only human motions without backgrounds are captured in the event images because of the property of event cameras. Furthermore, we have included visualized samples for all action classes in the dataset, which can be found in \figurename~\ref{dataset_vis_all}. There are a number of actions that involve the disappearing body parts at particular time instances, which can pose significant challenges to event-based human pose estimation.
We annotate human poses by using the annotation tool we designed, as shown in \figurename~\ref{annotation_tool}. The human keypoints are manually marked on RGB frames, while the keypoints are projected automatically to the event frames based on the calibrated parameters. 

With the annotation tool, we obtain all annotations of keypoints on event frames. \figurename~\ref{annotation_vis} lists some examples of event frames and their ground truth for human keypoints.}

\begin{figure}[ht]
\centering
\includegraphics[width=3.3in]{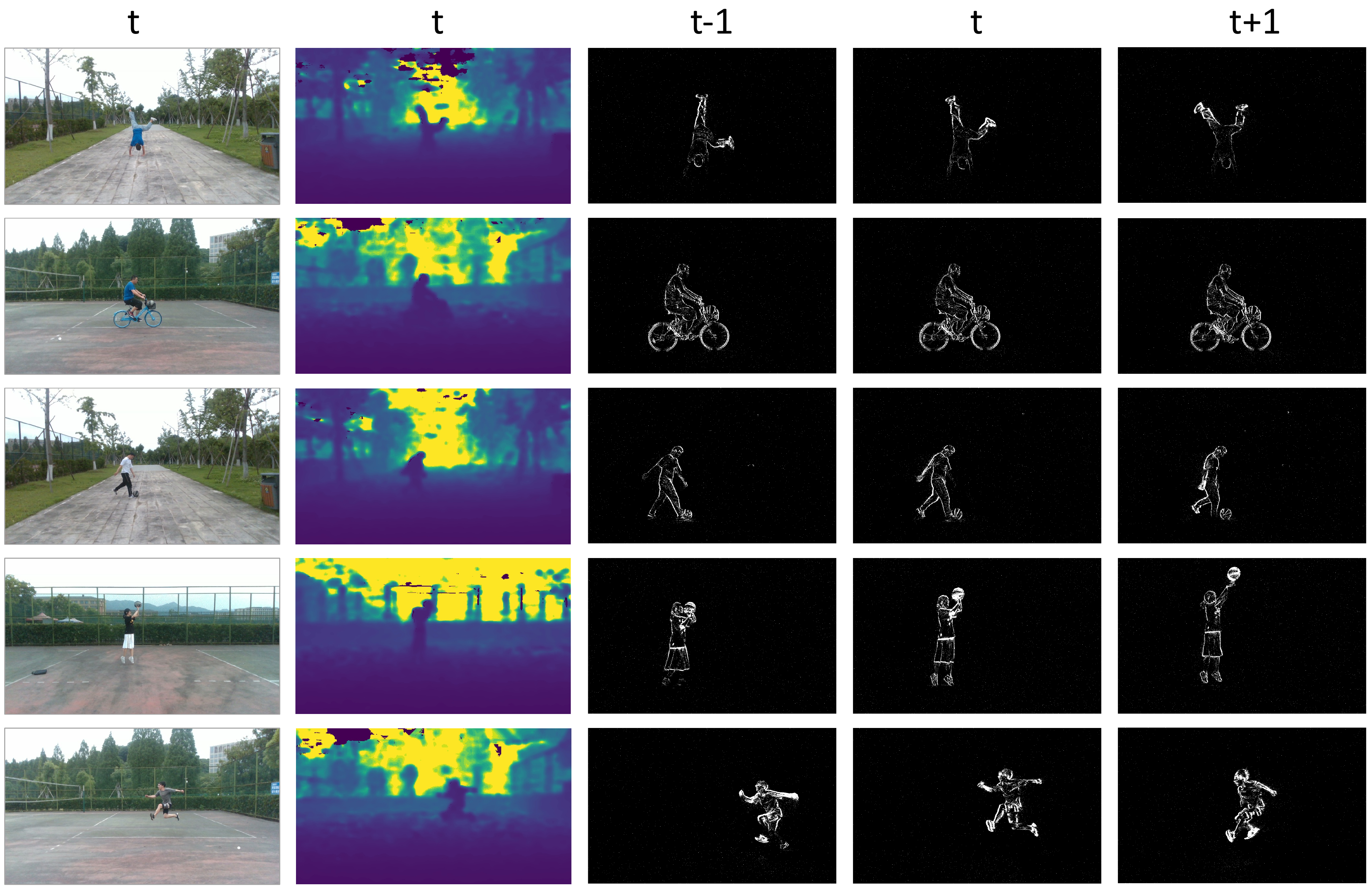}
\caption{Visualized examples of actions captured by the RGB, Depth, and Event Camera. Actions at the first column are RGB frames at $t$-th time, while actions at the rest of columns are the corresponding event frames and their neighboring frames.}
\label{dataset_vis}
\end{figure}

\begin{figure}[ht]
\centering
\includegraphics[width=2.5in]{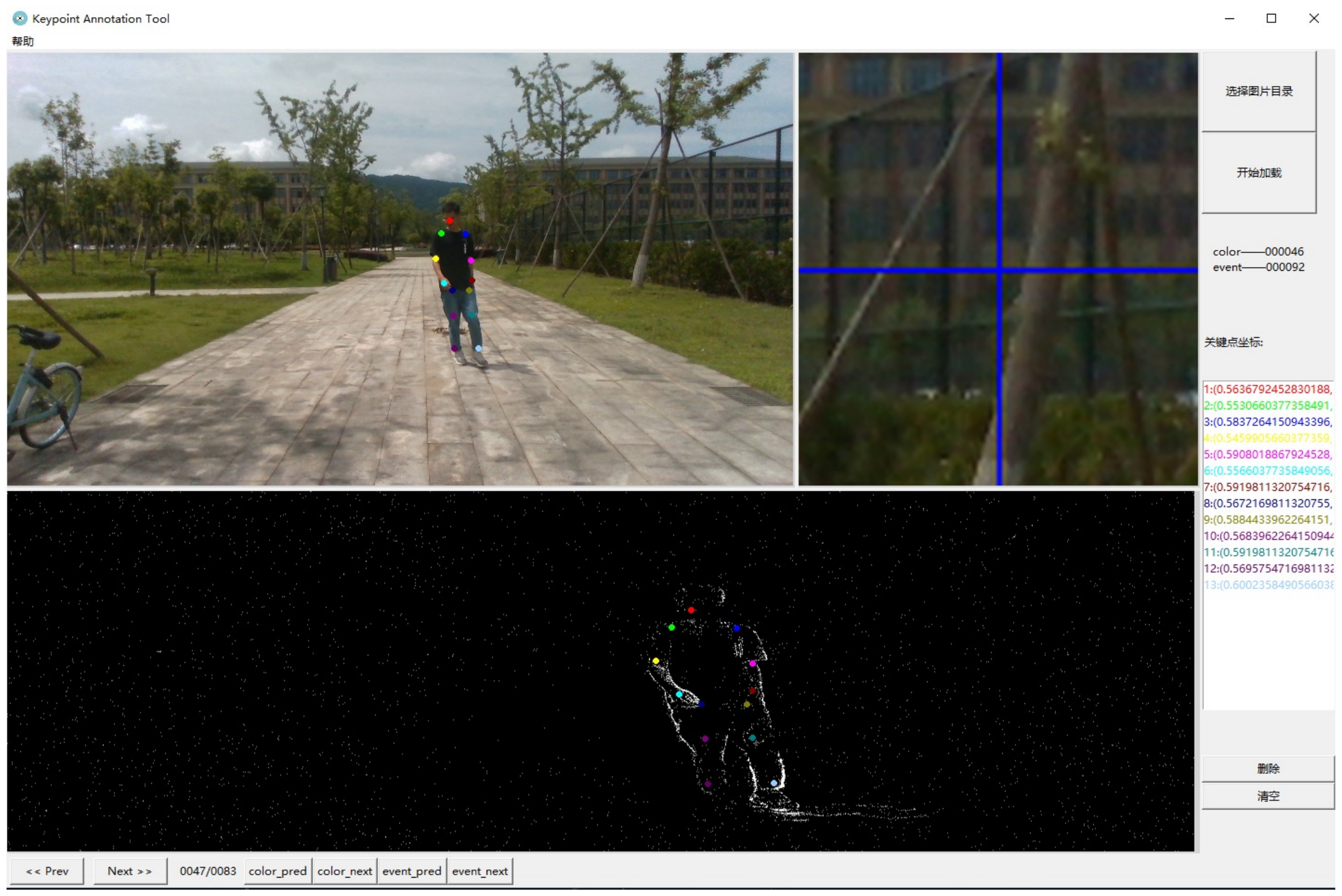}
\caption{The keypoint annotation tool we designed.}
\label{annotation_tool}
\end{figure} 

\begin{figure}[H]
\centering
\subfloat[]{\includegraphics[width=2.5in]{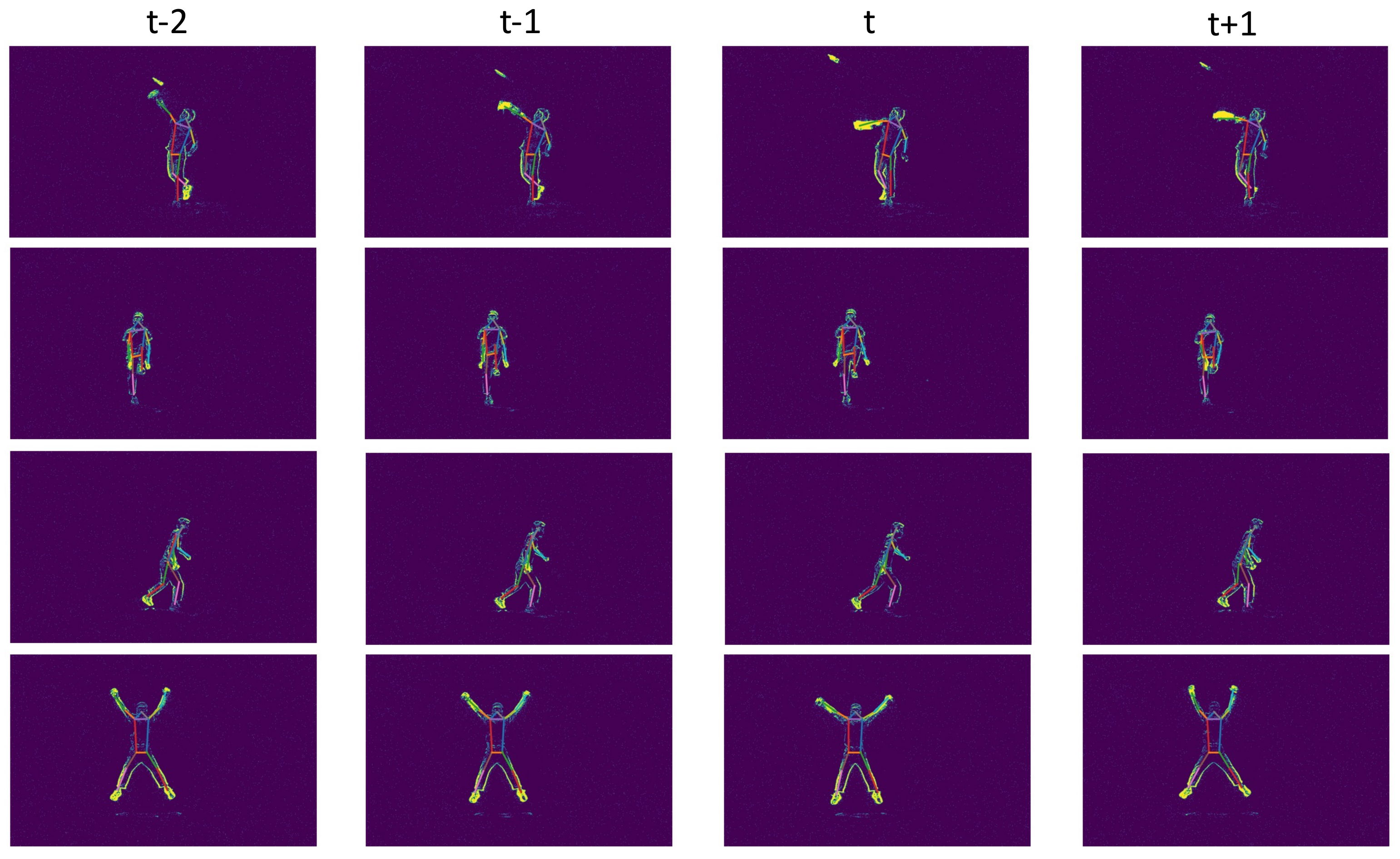}%
\label{app-fig4}}
\hfil
\subfloat[]{\includegraphics[width=2.5in]{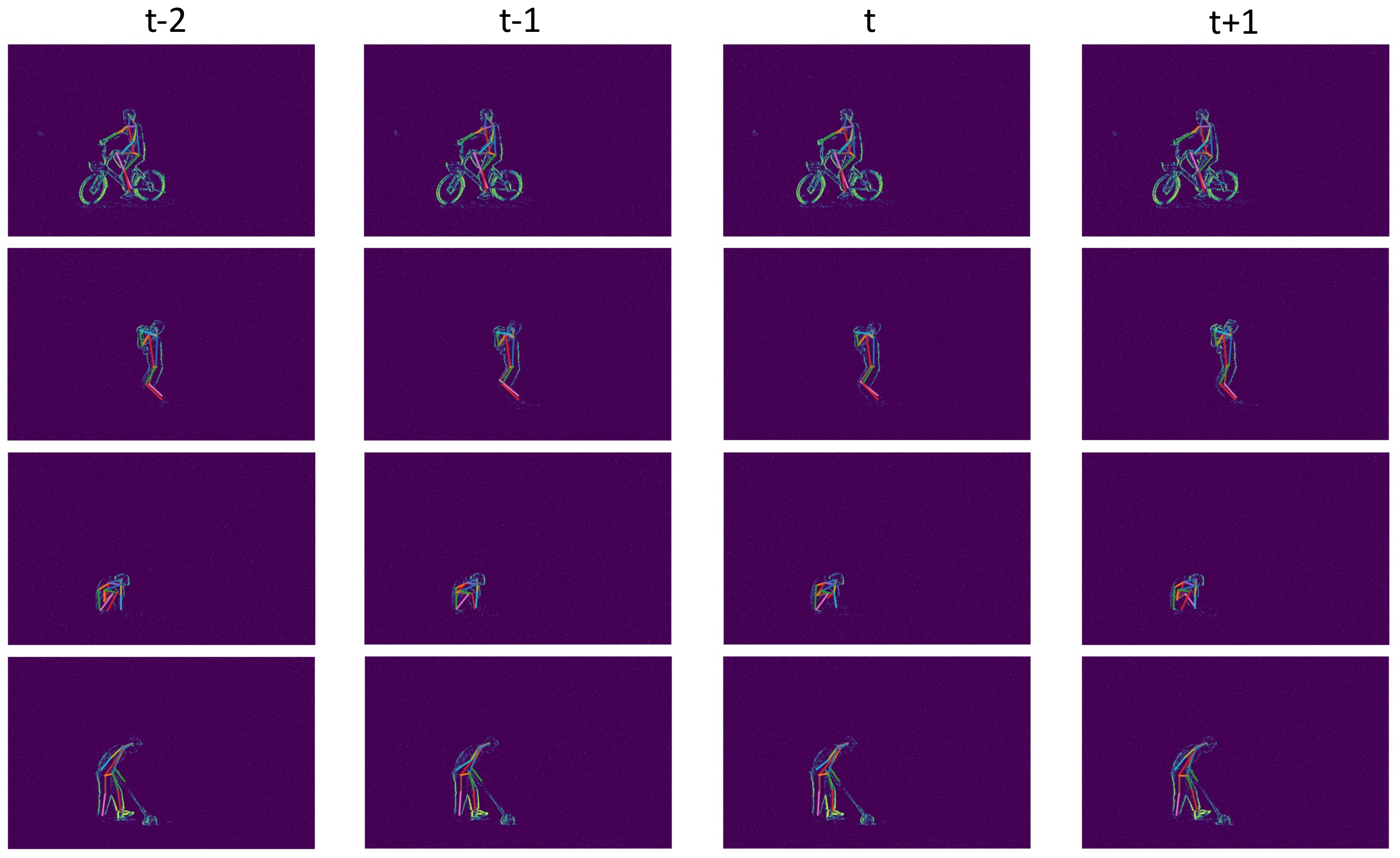}%
\label{app-fig5}}
\caption{Some annotated event-based action samples.}
\label{annotation_vis}
\end{figure}

\enlargethispage{-5in}

\end{document}